%% file: main.tex
\documentclass[twoside]{article}

\usepackage[accepted]{aistats2026}
\usepackage{subcaption}
\usepackage{xcolor}

\usepackage[utf8]{inputenc} 
\usepackage[T1]{fontenc}    
\usepackage{hyperref}       
\usepackage{url}            
\usepackage{booktabs}       
\usepackage{amsfonts}       
\usepackage{nicefrac}       
\usepackage{microtype}      
\usepackage{float}
\usepackage[ruled, vlined, linesnumbered]{algorithm2e}
\usepackage{amsmath}
\usepackage{graphicx}
\usepackage{subcaption}

\input{macros.tex}
\usepackage{wrapfig}
\usepackage[dvipsnames]{xcolor}
\usepackage[textsize=tiny]{todonotes}
\usepackage{xspace}
\usepackage{subcaption}
\usepackage{tikz}
\usetikzlibrary{arrows.meta,positioning,fit,calc,shadows.blur}
\usepackage{natbib}
\usepackage{algorithmic}
\usepackage{pifont}
\usepackage{enumitem}
\usepackage{titlesec}

\newif\iffinal

\iffinal
    \newcommand{\fix}[1]{}
    \newcommand{\yuxin}[1]{}
    \newcommand{\yuxinil}[1]{}
\else
    \newcommand{\fix}[1]{{\color{red} #1}}
    
    \newcommand{\yuxin}[1]{\todo[fancyline,color=NavyBlue!40]{Yuxin: #1}\xspace}

\fi

\newcommand{\algname}{\textsc{RLMesh}\xspace}

\begin{document}

\twocolumn[

\aistatstitle{
Accelerating PDE Surrogates via RL-Guided Mesh Optimization
}

%


\aistatsauthor{%
Yang Meng \And
Ruoxi Jiang \And
Zhuokai Zhao \And
Chong Liu \And
Rebecca Willett \And
Yuxin Chen
}

\aistatsaddress{%
UChicago \And
Fudan University \And
UChicago \And
UAlbany \And
UChicago \And
UChicago
}]

\begin{abstract}
Deep surrogate models for parametric partial differential equations (PDEs) can deliver high-fidelity approximations but remain prohibitively data-hungry: training often requires thousands of fine-grid simulations, each incurring substantial computational cost. 
To address this challenge, we introduce \algname, 
an end-to-end framework for efficient surrogate training under limited simulation budget.
%
The key idea is to use reinforcement learning (RL) to adaptively allocate mesh grid points non-uniformly within each simulation domain, focusing numerical resolution in regions most critical for accurate PDE solutions.
A lightweight proxy model further accelerates RL training by providing efficient reward estimates without full surrogate retraining. 
Experiments on PDE benchmarks 
demonstrate that \algname achieves competitive accuracy to baselines but with substantially fewer simulation queries. 
%
These results show that solver-level spatial adaptivity can dramatically improve the efficiency of surrogate training pipelines, enabling practical deployment of learning-based PDE surrogates across a wide range of problems.
\end{abstract}

\input{sections/introduction}

\input{sections/related_work}
\input{sections/problem_formulation}
\input{sections/method}

\input{sections/experiments}
\input{sections/analysis}

\input{sections/conclusion}

\input{sections/acknowledgment}

\bibliographystyle{plainnat}
\bibliography{reference}

\newpage
\appendix
\thispagestyle{empty}

\onecolumn
\aistatstitle{Accelerating PDE Surrogates via RL-Guided Mesh Optimization: Supplementary Materials}

\section{Details on Experiment Setup}

\subsection{Fourier Neural Operator}\label{sec:fno-supp}

In this section, we describe the Fourier Neural Operator (FNO) architecture used as our surrogate model, detailing the architectural modifications made to handle non-uniform grids and the training procedures employed in both pretraining and active learning phases.

\subsubsection{Modification}
The standard FNO operates efficiently on uniform grids by learning global convolutions in the frequency domain. However, our framework requires flexible modeling on non-uniform grid points placements. To this end, we propose architectural modifications and extensions to enable robust learning from non-uniformly distributed grid points.

\begin{itemize}
    \item \textbf{Fourier feature encoding:} We augment raw spatial coordinates with a learnable Fourier feature embedding. This encoding transforms each coordinate into a higher-dimensional space using sinusoidal functions with Gaussian-sampled frequencies, enriching positional information for better generalization on irregular layouts.
    
    \item \textbf{Batch normalization on Fourier features:} To stabilize training and improve convergence, batch normalization is applied after the Fourier feature encoding. This normalization operates across feature channels and helps regularize the model when processing diverse input patterns.
    
    \item \textbf{Increased number of Fourier modes:} We increase the number of Fourier modes from typical values (e.g., 16) to 49 to capture finer spectral details relevant to the 129-point grid resolution, enhancing the model's capacity to learn complex solution structures.
    
    \item \textbf{Multi-head self-attention with residual connection:} A multi-head self-attention layer is integrated into the spectral convolution stack to further model long-range dependencies across locations. Combined with residual connections, this addition strengthens global feature interactions, which are critical for capturing the complex spatial correlations present in non-uniform mesh grid points data.
\end{itemize}

Together, these modifications enable the surrogate $\surrop$ to learn PDE solution operators from irregular inputs, improving predictive robustness in the active learning loop.

\subsubsection{Training Details}
We pretrained an ensemble of five FNOs on a subset of 100 samples from the chosen-fidelity (60-point) training data for stability and robustness. Each FNO was configured with 49 Fourier modes and a width of 64 channels, incorporating our proposed modifications including Fourier feature encoding with batch normalization and a multi-head self-attention layer with residual connections. Models were trained with a negative log-likelihood (NLL) loss over 250 epochs (batch size 32) using Adam (learning rate \(1\!\times\!10^{-3}\)). The NLL objective allows each model to capture predictive uncertainty by learning both the mean and variance; at inference we aggregate ensemble predictions by averaging means and, when needed, combining variances.

In the active learning phase, grid-point placements are iteratively selected by a reinforcement learning policy, with batches of 50 newly selected samples per iteration across 18 iterations (total 900 samples). After each batch, all five FNOs are retrained on the union of the pretraining set and the newly collected samples using the same hyperparameters. This retraining strategy lets the surrogate ensemble progressively refine accuracy in regions deemed most informative by the RL policy. Concurrently, the RL agent's policy network is updated from rollout data, improving grid-point placement over time to minimize reconstruction error.

In both phases, we train the FNO surrogate on non-uniform samples uses a mask-based set-to-grid interface. Sparse observations queried at non-uniform mesh points are mapped to the nearest grid locations; a binary mask records which grid points are observed. During training, the FNO predicts the full field on the standard grid, but the loss is computed only at masked locations. This preserves the standard FNO architecture and spectral convolutions, while enabling learning from irregularly sampled data.


\subsection{Reinforcement Learning Agent}

In this section, we present the RL component of our framework, detailing the Deep Q-Network (DQN) architecture, the offline pretraining procedure using oracle demonstrations, and the online training process through environment interaction and solver feedback.

\subsubsection{Deep Q-Network}

We implement a DQN agent, which uses a neural network function approximator to estimate the Q-values for each state-action pair. The network architecture includes two hidden layers with ReLU activations and a final Tanh layer bounding Q-values. To stabilize training, we employ an experience replay buffer that stores transitions, enabling off-policy learning through mini-batch updates, and maintain a separate target network updated periodically to reduce oscillations and divergence. Exploration is governed via an epsilon-greedy policy, with epsilon annealed from an initial value to a minimum threshold to balance exploration and exploitation. The DQN agent learns from both intermediate discounted rewards and terminal rewards, updating its Q-values through gradient descent minimizing temporal difference errors.

\subsubsection{Offline Pretraining} 
\label{sec:rl-pretrain-details}

To initialize the reinforcement learning agent effectively, we perform offline pretraining using expert demonstrations derived from a heuristic policy informed by domain knowledge of the PDE. This expert policy leverages oracle information about the physical characteristics of sharp fronts and informative points, ensuring selection of grid points with high utility. The oracle follows a simple rule that combines local variation and spatial coverage: it prioritizes regions with strong gradients or curvature, while ensuring that the selected mesh grid points are spread evenly across the domain. The same principle is applied consistently to all problems. We collect transitions, specifically state-action pairs, over 100 training instances using this policy. The transitions are stored as tuples of the environment state and the chosen action, capturing the expert's behavior without requiring rewards or next states. Subsequently, imitation learning is applied to pretrain the RL agent by treating this offline dataset as supervised learning examples, where the agent's Q-function approximator is trained to predict expert actions given states. For the DQN agent, this involves training the neural network policy to minimize cross-entropy loss between predicted action logits and expert actions. This approach accelerates convergence during online learning by providing a strong initial policy derived from domain expertise. The imitation dataset and pretraining process are logged extensively to ensure reproducibility and monitor training progress.

\subsubsection{Online Training} 
During the online phase, the RL agent interacts with the environment to sequentially select grid points within each episode. After an episode concludes, the selected grid points define a nonuniform mesh that is passed to the numerical solver to obtain the PDE solution restricted to those chosen locations. This newly generated data is then used to retrain the FNO surrogate model. The terminal reward for the RL agent is computed based on the improvement in the proxy model's predictive performance after retraining (the exact reward formulation is described in the next section).  We train the DQN with Adam (learning rate \(1\times10^{-4}\), weight decay \(1\times10^{-4}\)), discount \(\gamma=0.99\), a replay buffer of 10{,}000, batch size 64, \(\epsilon\)-greedy exploration decaying from 1.0 to 0.1 at 0.995 per step, target-network synchronization every 100 updates, and gradient clipping at 1.0. Meanwhile, the RL policy network is updated using Monte Carlo backups of discounted rewards, allowing the agent to gradually learn more effective and informative mesh grid points placement strategies over time.



\subsection{Proxy Model} \label{sec:proxy}
In this section, we motivate and select a lightweight proxy to stand in for the FNO, describe how the proxy is trained and updated online, and formalize the terminal-reward computation used by the RL agent.

\subsubsection{Alignment between FNO and Proxy Models}
\begin{figure}[!htbp]
  \centering
  \captionsetup[subfigure]{justification=centering}

  \begin{subfigure}[t]{0.6\linewidth}
    \centering
    \includegraphics[width=\linewidth]{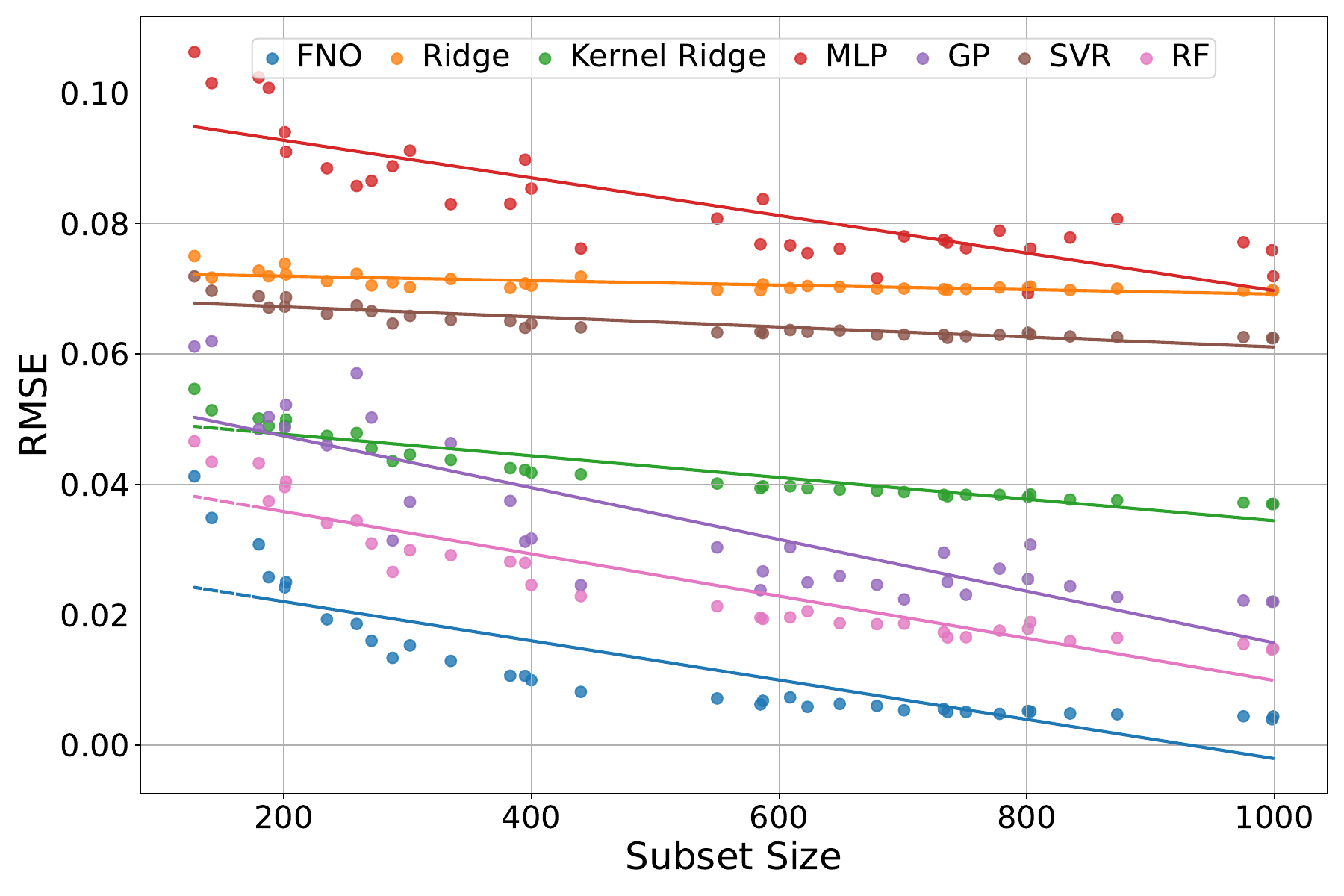}
    \caption{Model correlation}
    \label{fig:proxy-correlation}
  \end{subfigure}

  \vspace{0.6em} 

  \begin{subfigure}[t]{\linewidth}
    \centering
    \includegraphics[width=\linewidth]{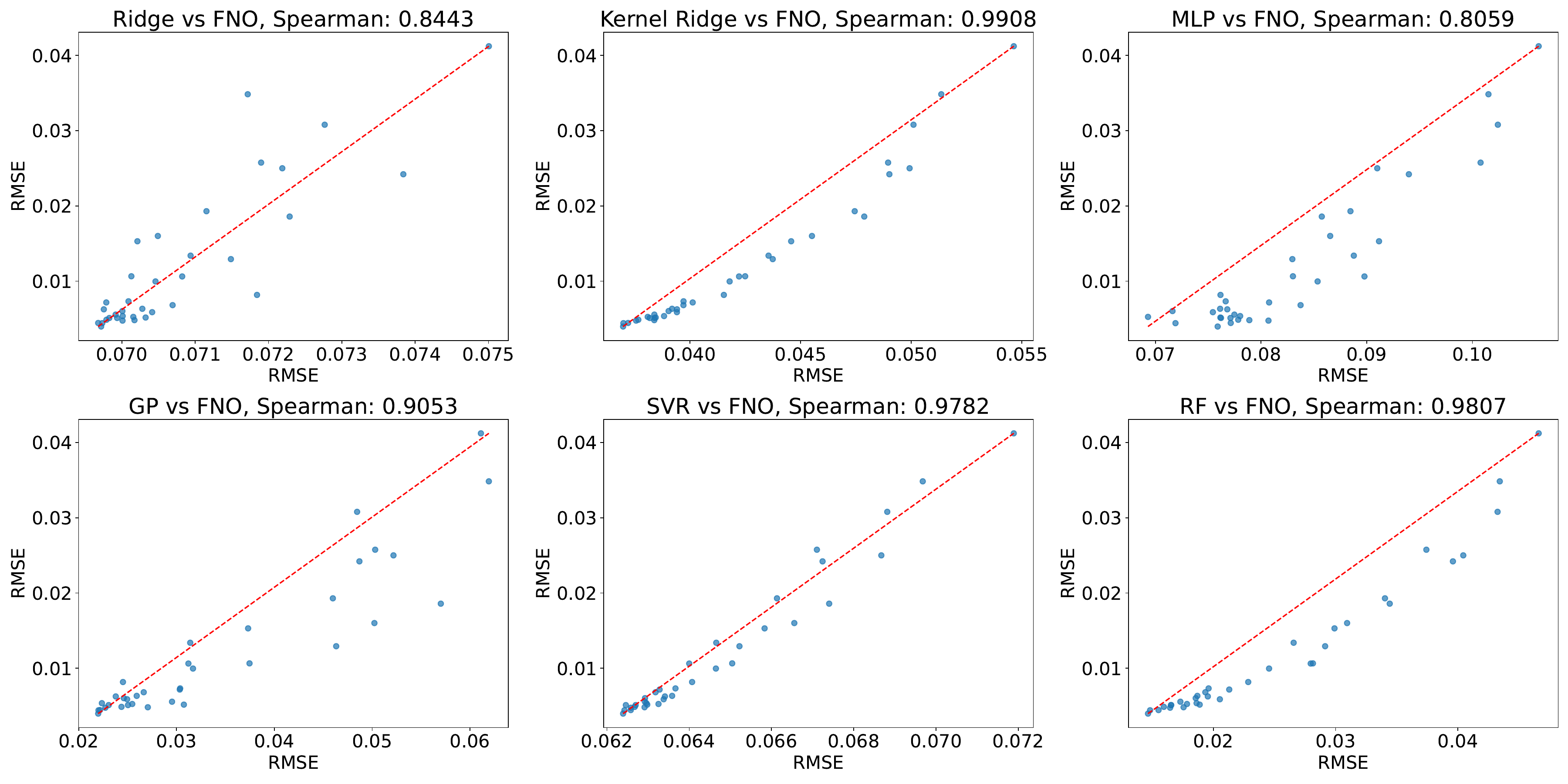}
    \caption{Model RMSE}
    \label{fig:proxy-rmse}
  \end{subfigure}
  \caption{Model choice assessment for the proxy. (a) Correlation between the proxy and FNO across training subsets. (b) RMSE for candidate models across subset sizes.}
  \label{fig:proxy-choice}
\end{figure}

Intuitively, using the surrogate's downstream error as the RL signal aligns the agent's objective with the end task—accurate PDE prediction—so each placement is rewarded only insofar as it improves real generalization. However, evaluating an FNO ensemble per episode is prohibitively slow, so we introduce a lightweight proxy to stand in for the FNO when computing terminal rewards. Figure~\ref{fig:proxy-choice} compares candidate proxies trained across varying subset sizes and assesses their agreement with the FNO: panel (a) shows RMSE trends as data grow; panel (b) plots proxy versus FNO errors with monotonicity (Spearman) scores. Kernel Ridge Regression (RBF) exhibits the strongest and most stable alignment with the FNO (near-perfect monotonic correlation) while remaining orders of magnitude faster than retraining the ensemble. We therefore adopt Kernel Ridge as the proxy for reward computation: after each episode's solver call, we append the new sample, retrain the proxy in milliseconds, and compute the terminal reward from the proxy's test-RMSE improvement (exact reward form in the next section).

\subsubsection{Training Details}
To accelerate reward calculation, we trained the kernel ridge regression proxy on an initial subset of 100 labeled training instances sampled from the full 129-point fidelity data. The kernel ridge model uses a Gaussian radial basis function kernel with parameters $\alpha=0.1$ and $\gamma=1.0$ learned via cross-validation. The model is retrained online at each active learning episode, incorporating newly collected samples. We resample the non-uniform grid point observations onto a uniform grid via linear interpolation, since the kernel-ridge proxy assumes fixed-length inputs on a uniform lattice. This retraining typically completes in less than 0.8 seconds on an AWS EC2 g6e.xlarge instance equipped with an NVIDIA L40S GPU, enabling rapid updates to the proxy's predictive accuracy.

\subsubsection{Reward Computation for RL}

The RL reward $\reward$ is computed solely from the terminal improvement in $\proxy$ after retraining.  
Specifically, the raw reward for an episode is
\[
\reward_{\text{raw}} = -\kappa \bigl( \varepsilon_{\text{new}} - \varepsilon_{\text{old}} \bigr),
\]
where $\varepsilon_{\text{old}}$ and $\varepsilon_{\text{new}}$ are proxy RMSE values on a held-out test set before and after adding the newly acquired samples. $\kappa$ is a scaling constant and was set to $\kappa = 10^4$ for all experiments considered in this paper. To stabilize learning, we apply a piecewise scaling function to map $\reward_{\text{raw}}$ into $[-1,1]$:
\[
\text{scale}(\reward_{\text{raw}}) = 
\begin{cases}
0.8 \, \text{sgn}(\reward_{\text{raw}}), & |\reward_{\text{raw}}| < 0.01, \\
\left(0.8 + 0.2 \tfrac{|\reward_{\text{raw}}|}{0.1} \right) \text{sgn}(x), & 0.01 \leq |\reward_{\text{raw}}| < 0.1, \\
\reward_{\text{raw}}, & 0.1 \leq |\reward_{\text{raw}}| < 1, \\
\left(1.0 - 0.01 \tfrac{|\reward_{\text{raw}}|-1}{9} \right) \text{sgn}(\reward_{\text{raw}}), & 1 \leq |\reward_{\text{raw}}| < 10, \\
\text{sgn}(\reward_{\text{raw}}) \min\!\left\{1.0, \, 0.99 + 0.01 \ln \tfrac{|\reward_{\text{raw}}|}{10} \right\}, & |\reward_{\text{raw}}| \geq 10.
\end{cases}
\]

No rewards are provided at intermediate steps; the entire episode's credit is assigned via this terminal proxy-improvement signal. We avoid intermediate heuristics because we find no reliable per-step surrogate that aligns with downstream generalization, and such signals can introduce bias toward myopic or miscalibrated placements.

\subsection{Experiment Details}\label{sec:exp-supp}

We evaluate on three PDE families—1D Burgers (terminal-time prediction), 2D Darcy (steady mapping), and Lorenz-96 system (chaotic spatiotemporal dynamics)—with 1{,}000 training and 200 held-out test instances per problem. Of the 1{,}000 training instances, 100 are used for surrogate pretraining and the remainder for active acquisition. We sweep per-instance budgets and report test RMSE on a fixed dense grid (129 points for Burgers; 128$\times$128 for Darcy) as a function of labeled samples and wall-clock cost. Unless noted, all hyperparameters (RL agent, proxy, surrogate training schedule) are fixed across runs, and all methods (RL and baselines: uniform, random, gradient, variance, intensity) share the same sbudget, number of solver queries per iteration, and surrogate retraining cadence for fairness. Results are averaged over five independent seeds (mean~$\pm$~std) with a fixed train/test split per seed. Experiments were run on Amazon EC2 instances and an HPC cluster with identical code, data, and hyperparameters using L40S GPU. We get our data from the PDEBench dataset under CC-BY-4.0~\citep{Takamoto2022PDEBench}.
We additionally evaluate RLMesh on the Lorenz--96 system, a 1D lattice dynamical model with $N=60$ state variables and periodic coupling. Each instance is generated by integrating the system from $t=0$ to $t=1$ with step size $\Delta t=0.01$, and forming a supervised pair $(x(0), x(1))$. We impose a strict per-instance budget of 15 queried coordinates (25\% of the state). Results are averaged over five independent seeds, and all baselines share the same query budget and acquisition protocol.


\subsection{PDE Simulator}\label{sec:exp-simulator}
For Burgers we use a custom finite-volume solver with Dirichlet walls at the domain endpoints, integrated in time with SciPy's odeint (LSODA) under tight tolerances. Spatially, we employ MUSCL (Van Leer) reconstruction and a blended convective flux that interpolates between Godunov (sharper, less diffusive) and Rusanov (more robust); viscosity is handled via a face-based diffusive flux. The solver supports uniform and non-uniform grids: on non-uniform meshes we reconstruct left/right face states using node-to-face distances and compute face spacings explicitly in both convection and diffusion; boundary values are pinned only if boundary nodes are present, so interior-only sets are respected. To stabilize pathological layouts from sparse mesh grid points, we add a geometry augmentation step that inserts virtual midpoints to cap the maximum gap (relative to the nominal spacing) and the adjacent-gap ratio, with optional wall anchoring; the initial condition is re-interpolated onto the augmented grid before integration. We use a simple CFL-based heuristic for the integrator's initial step and provide an option to restrict outputs back to the original set for consistent evaluation. Consequently, runtime is not monotone in the number of mesh grid points: very sparse, irregular 20-point layouts can trigger more geometry augmentation and force smaller stable time steps than a better-distributed 40-point layout, so the 20-point solve can actually take longer (see Fig. \ref{fig:tradeoff}). Likewise, the more-point case can benefit from vectorized operations and larger stable steps. 

We quantitatively validate our non-uniform finite-volume solver against dense uniform-grid oracle solutions, obtaining MAE $7.42\times10^{-3}\pm2.91\times10^{-3}$ and RMSE $1.60\times10^{-2}$ under the sparse regimes used by RLMesh. While developing a production-grade, fully verified adaptive solver integrated with industrial PDE frameworks is beyond the scope of this work, these checks and the use of a common oracle across all comparisons ensure that solver accuracy is sufficient and does not affect soundness or fairness.

Extending the time–error tradeoff analysis to 2D Darcy is an interesting direction for future work. However, constructing a robust non-uniform 2D Darcy solver involves substantial numerical complexity, including geometry-aware meshing, stability constraints, and topology-dependent stencil construction, which are typically handled by specialized AMR frameworks. Developing such solvers is beyond the scope of this proof-of-concept study, which focuses on demonstrating the feasibility of the proposed approach.

\section{Additional Experimental Results}

\subsection{Fewer Grid Points Performance} \label{sec:fewergrid-exp}

\begin{figure}[t]
    \centering
    \includegraphics[width=0.5\linewidth]{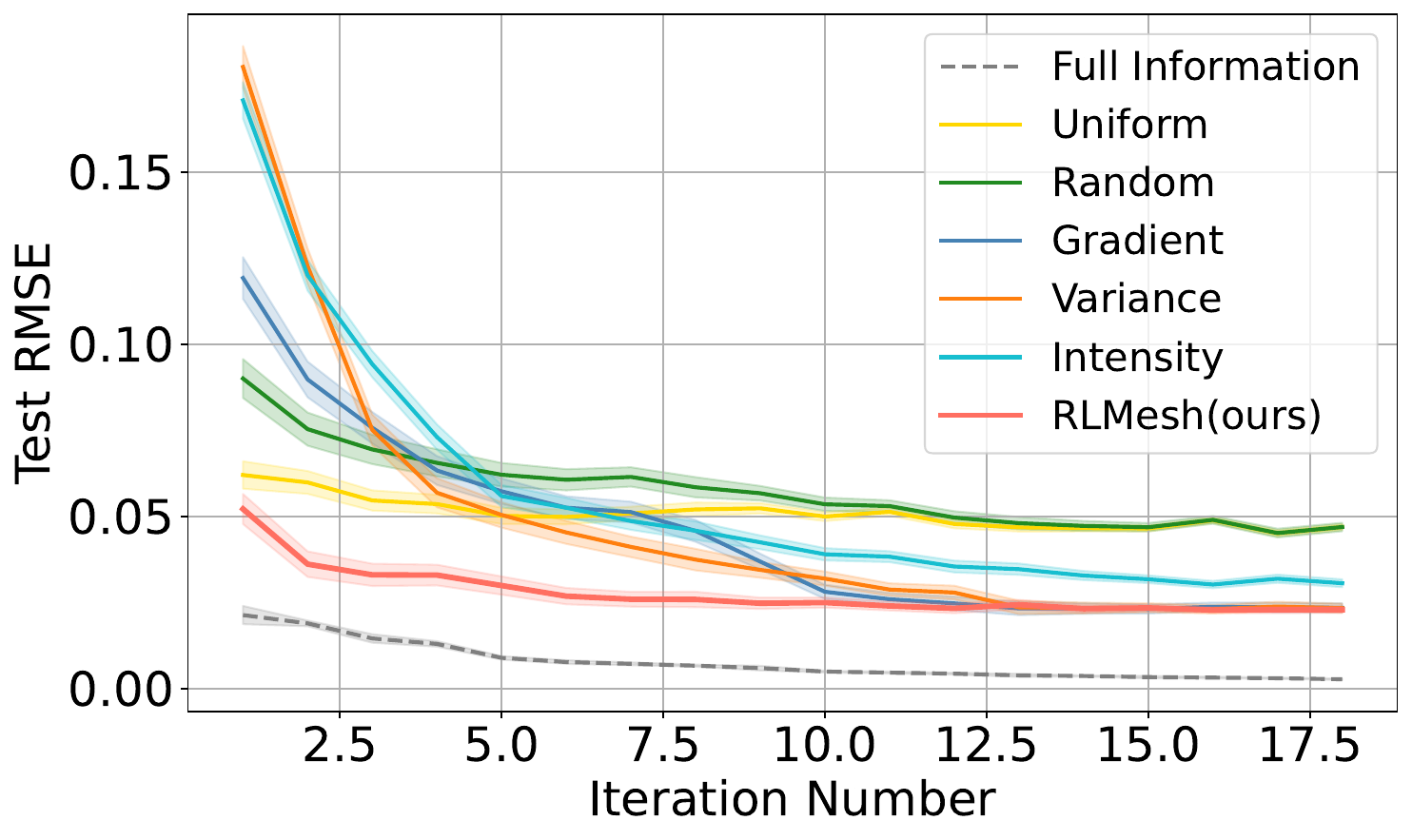}
    \caption{Active learning performance on Burgers with 20 per-instance budget}
    \label{fig:20_pts}
\end{figure}

In \figref{fig:20_pts}, all baselines degrade under the smaller 20-point budget: their curves fall more slowly and plateau higher than with larger budgets. RLMesh (ours) still outperforms the heuristics across iterations, but the gap to the full-information oracle widens (same for all other baselines), indicating that limited sensing amplifies the difficulty. This pattern is consistent with a capacity/conditioning bottleneck in the surrogate (FNO) under sparse measurements rather than a failure of the placement policy.

\subsection{Simulator Performance}\label{sec:app:custom_simulator}
\begin{figure}[!htbp]
    \centering
    \includegraphics[width=0.5\linewidth]{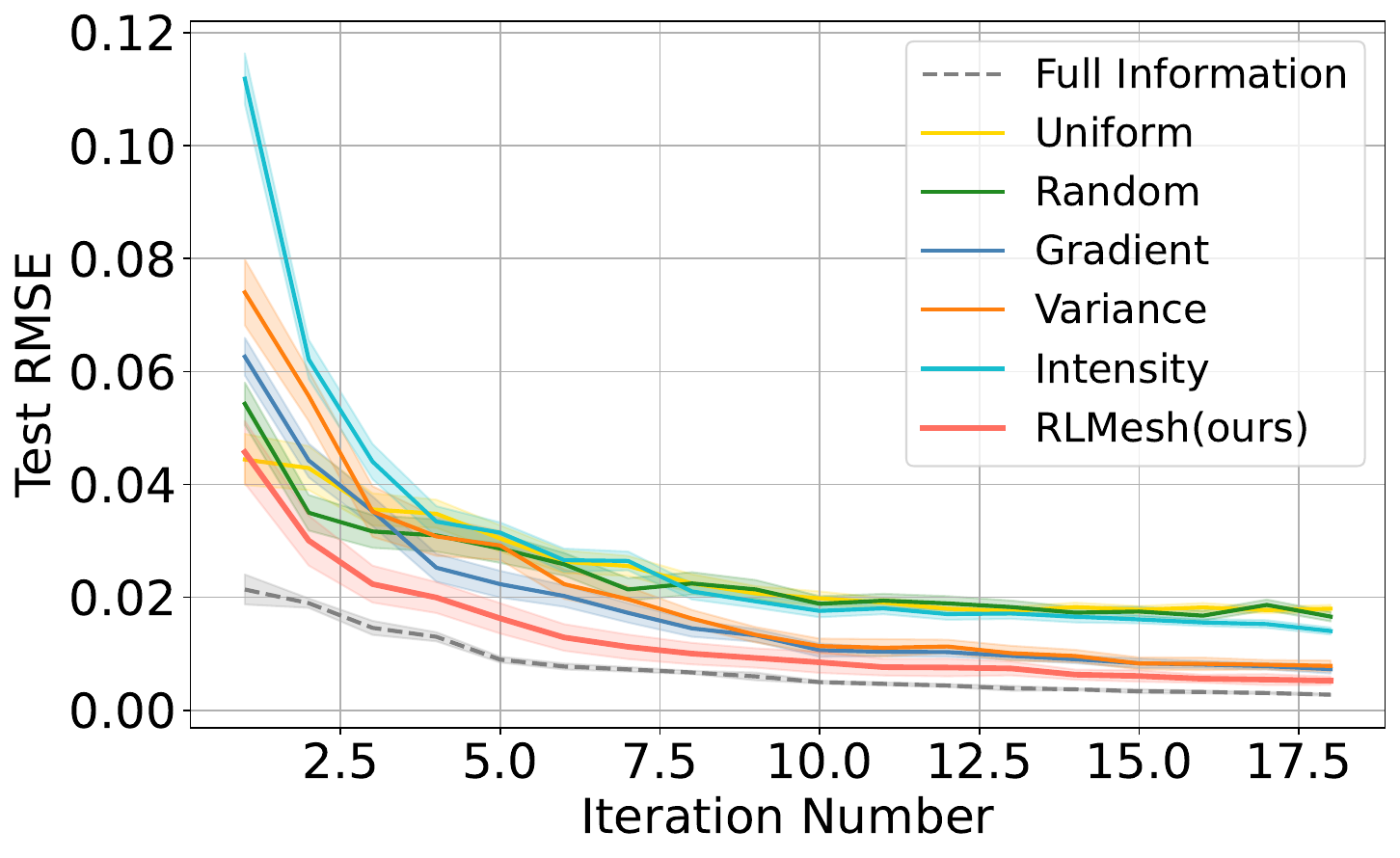}
    \vspace{-0.05in}
    \caption{Active learning performance on Burgers with PDE simulator. \algname consistently outperforms baselines under identical budgets.}
    \label{fig:with_sim}
    \vspace{-0.1in}
\end{figure}

\figref{fig:with_sim} shows that with the simulator enabled, the results provide the same insight as before: \algname remains the best non-oracle method across iterations.




\subsection{Mesh Grid Illustrations}

\begin{figure}
    \centering
    \includegraphics[width=1\linewidth]{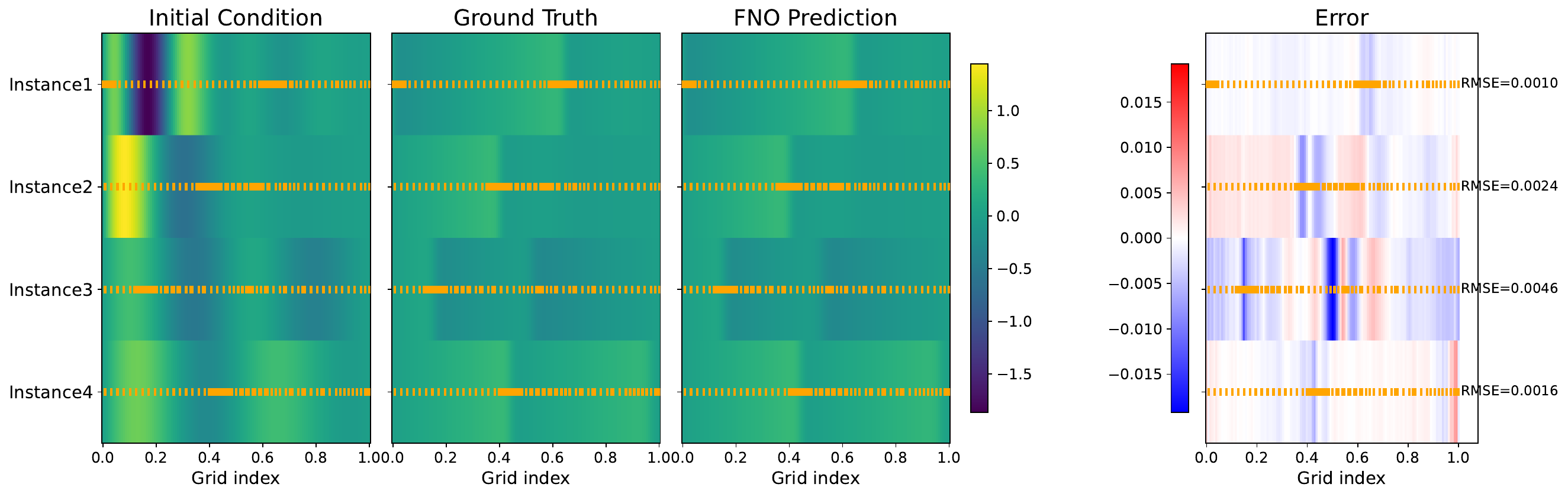}
    \caption{Burgers prediction with and without mesh overlay}
    \label{fig:burger_4}
\end{figure}

\figref{fig:burger_4} is a joint visualization designed to reveal the relationship between our selected grid points and model errors. It shows four Burgers instances (rows) across four views (columns): the initial condition, the ground-truth solution, the FNO prediction, and the error (prediction minus ground truth). The error panel uses a symmetric blue–white–red scale, with blue indicating under-prediction and red indicating over-prediction. The horizontal axis in every panel is the normalized spatial grid. Vertical orange tick marks denote the chosen mesh points; the same set is overlaid on all four views for each row to show how measurements align with field features. Per-instance RMSE values are listed on the right side of the error column. Overall, the placements concentrate near steep transitions and high-error bands, indicating that our selection strategy effectively targets the regions where the model struggles (the same pattern appears in the Darcy counterpart).

\figref{fig:darcy_4} is a joint visualization built to show how our selected grid points relate to model errors. Each Darcy instance appears twice—first without the overlay and then with the same overlay—to compare the fields and the sampling pattern. Within each row, the four panels (left to right) show the initial condition, the ground-truth solution, the FNO prediction, and the error (prediction minus ground truth) rendered with a symmetric blue–white–red scale; per-instance RMSE is annotated on the error panel. The first three panels share a common color range so magnitudes are directly comparable. In the ``with mesh overlay'' rows, an orange triangulated overlay marks the chosen grid points and their connectivity across all three left panels. Across instances, these placements align with steep transitions and high-error bands, indicating that the selection strategy effectively targets the regions where the model is most challenged.


\begin{figure*}[t]
  \centering
  \captionsetup[subfigure]{justification=centering}

  \begin{subfigure}[t]{\linewidth}
    \centering
    \includegraphics[width=\linewidth]{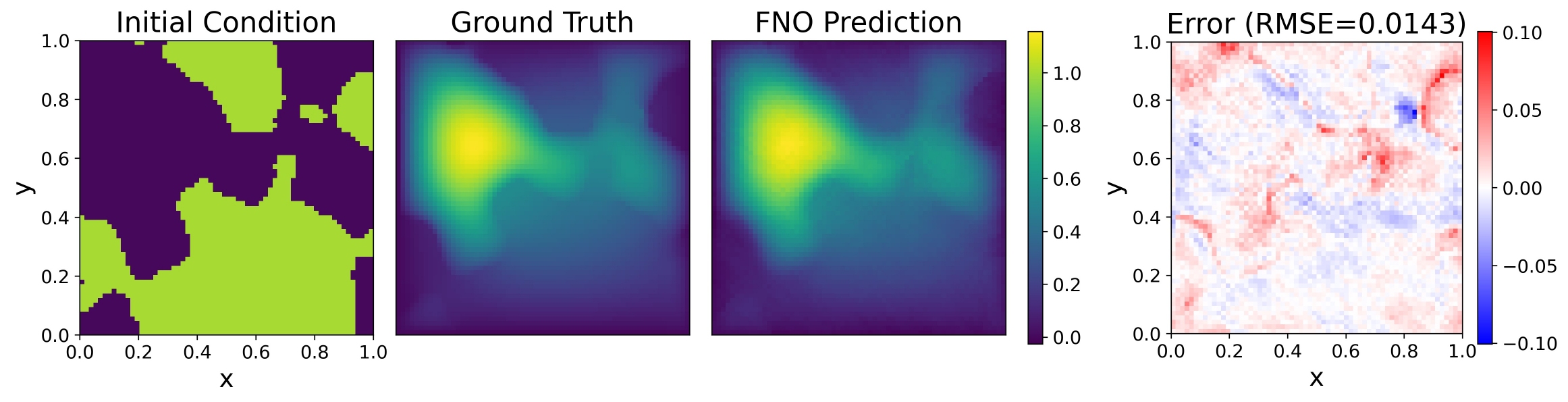}
    \caption{Instance 1 — no mesh}
    \label{fig:darcy_pred1_new}
  \end{subfigure}

  \vspace{0.6em}

  \begin{subfigure}[t]{\linewidth}
    \centering
    \includegraphics[width=\linewidth]{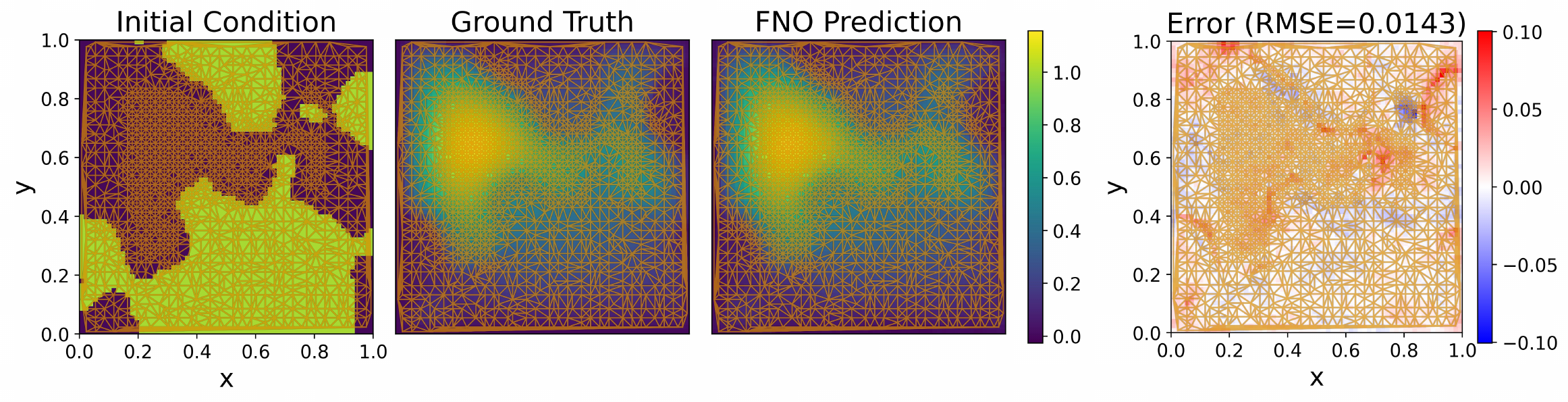}
    \caption{Instance 1 — with mesh overlay}
    \label{fig:darcy_pred1_mesh}
  \end{subfigure}

  \vspace{0.8em}

  \begin{subfigure}[t]{\linewidth}
    \centering
    \includegraphics[width=\linewidth]{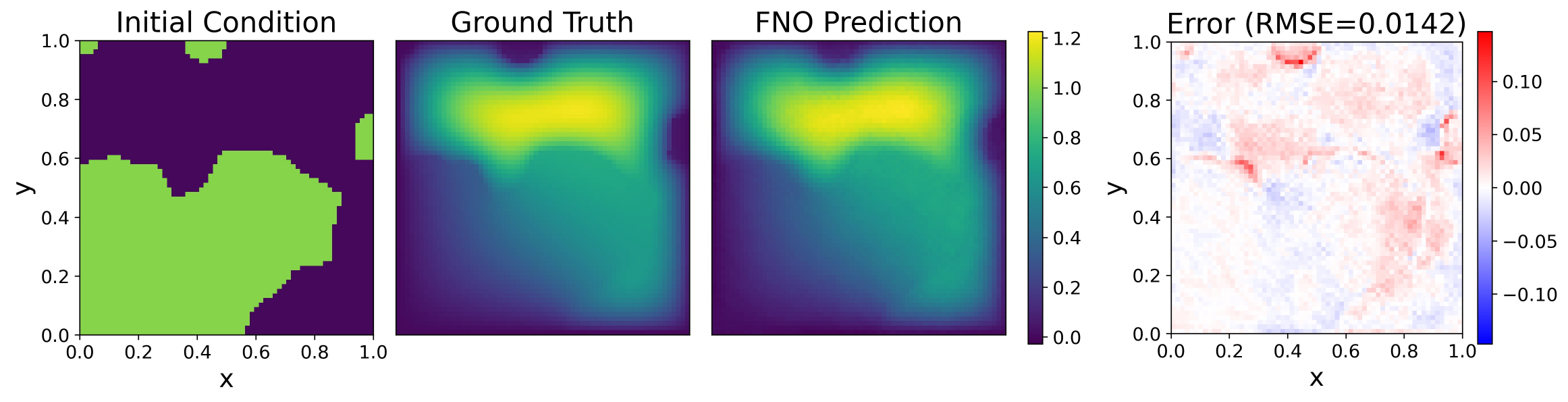}
    \caption{Instance 2 — no mesh}
    \label{fig:darcy_pred2_new}
  \end{subfigure}

  \vspace{0.6em}

  \begin{subfigure}[t]{\linewidth}
    \centering
    \includegraphics[width=\linewidth]{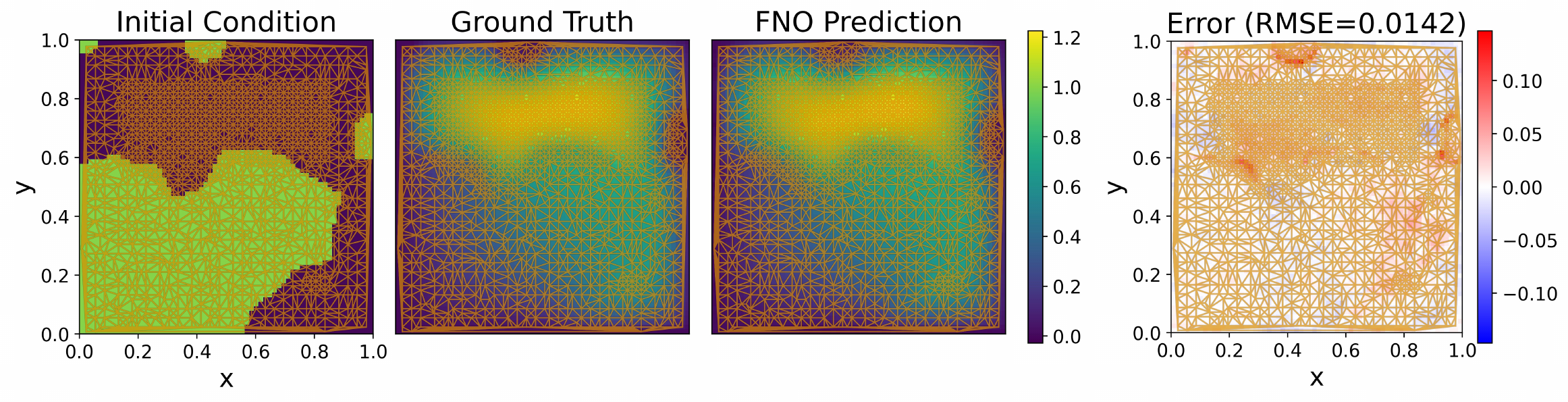}
    \caption{Instance 2 — with mesh overlay}
    \label{fig:darcy_pred2_mesh}
  \end{subfigure}

  \caption{Darcy prediction with and without mesh overlay (instances 1–2).}
  \label{fig:darcy_all}
\end{figure*}


\begin{figure*}[t]\ContinuedFloat
  \setcounter{subfigure}{4} 
  \centering
  \captionsetup[subfigure]{justification=centering}

  \begin{subfigure}[t]{\linewidth}
    \centering
    \includegraphics[width=\linewidth]{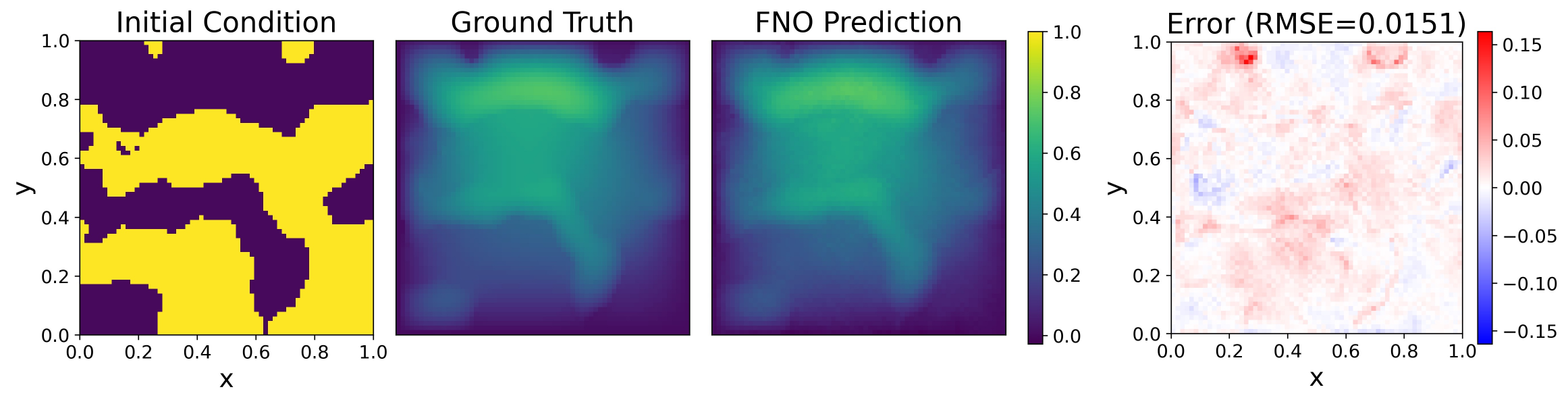}
    \caption{Instance 3 — no mesh}
    \label{fig:darcy_pred3_new}
  \end{subfigure}

  \vspace{0.6em}

  \begin{subfigure}[t]{\linewidth}
    \centering
    \includegraphics[width=\linewidth]{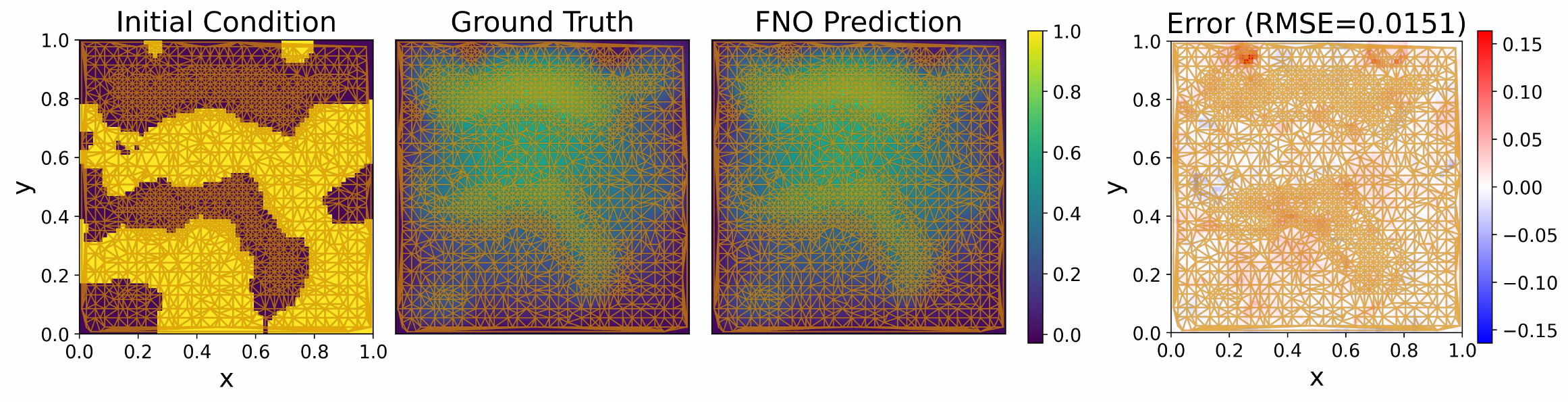}
    \caption{Instance 3 — with mesh overlay}
    \label{fig:darcy_pred3_mesh}
  \end{subfigure}

  \vspace{0.8em}

  \begin{subfigure}[t]{\linewidth}
    \centering
    \includegraphics[width=\linewidth]{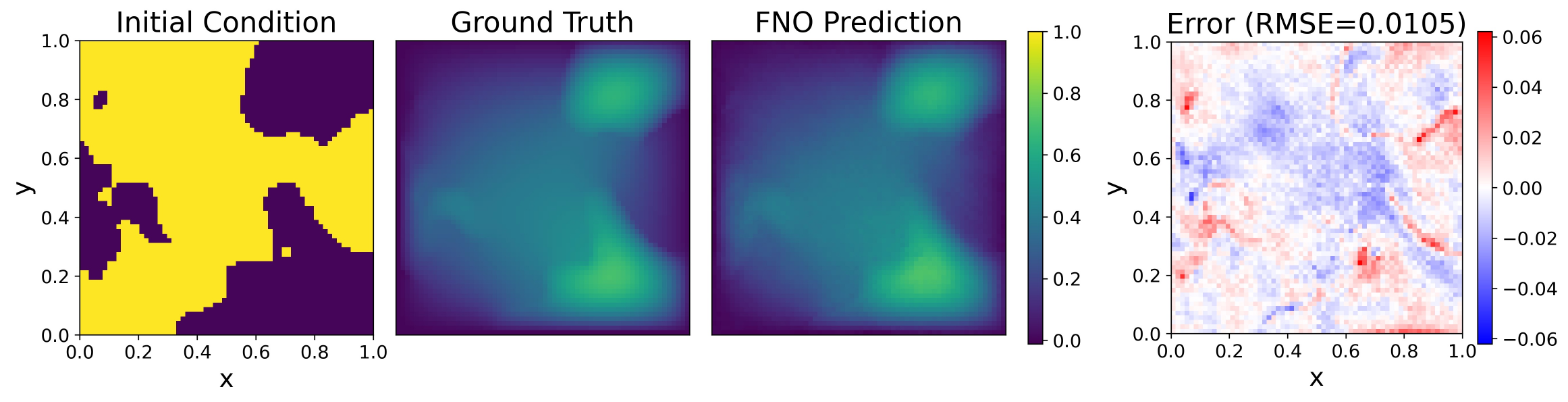}
    \caption{Instance 4 — no mesh}
    \label{fig:darcy_pred4_new}
  \end{subfigure}

  \vspace{0.6em}

  \begin{subfigure}[t]{\linewidth}
    \centering
    \includegraphics[width=\linewidth]{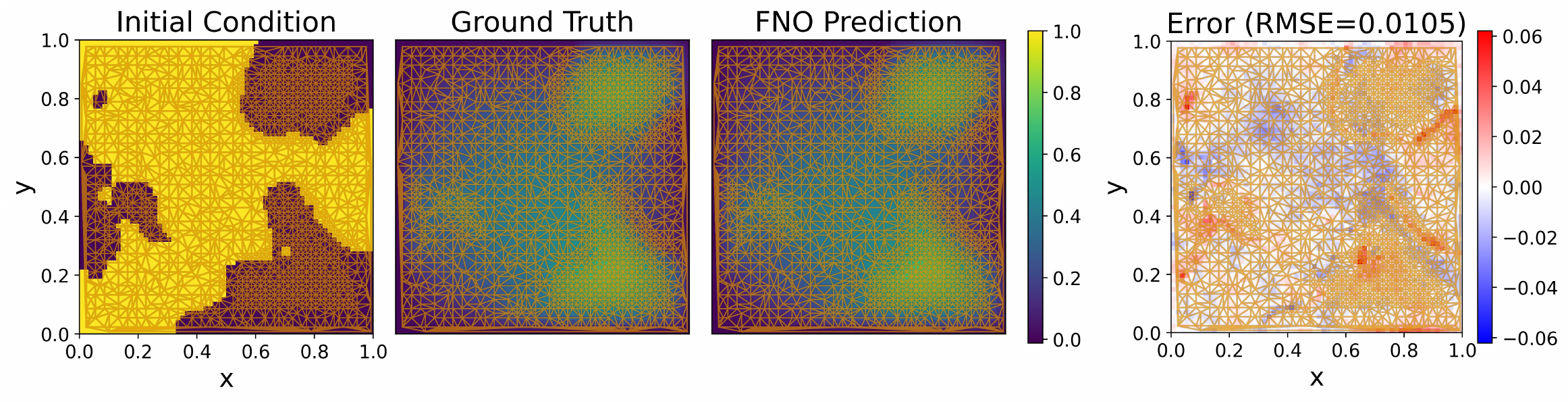}
    \caption{Instance 4 — with mesh overlay}
    \label{fig:darcy_pred4_mesh}
  \end{subfigure}

  \caption{Darcy predictions with and without mesh overlay (instances 3–4, continued).}
  \label{fig:darcy_4}
\end{figure*}

\begin{figure}[t]
  \centering
  \captionsetup[subfigure]{justification=centering}

  \begin{subfigure}[t]{0.485\linewidth}
    \centering
    \includegraphics[width=\linewidth]{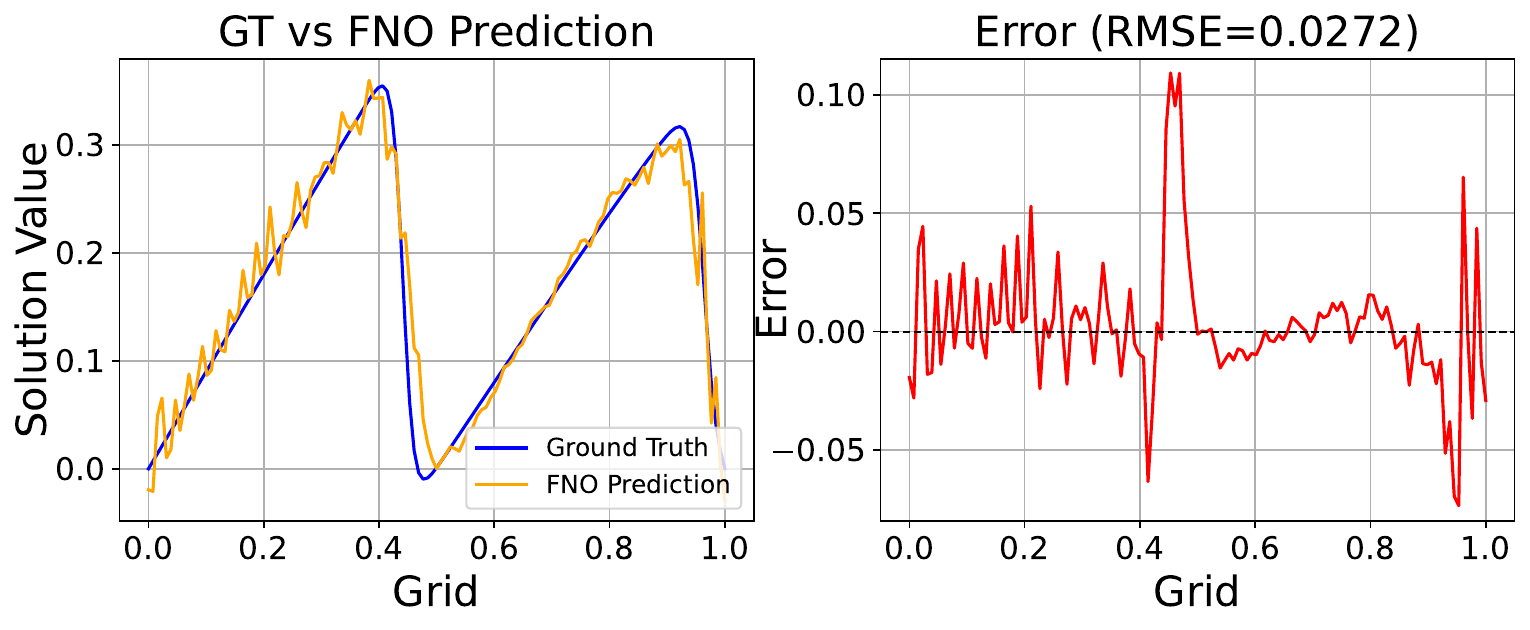}
    \caption{pretrained}
    \label{fig:burger_pred1}
  \end{subfigure}\hfill
  \begin{subfigure}[t]{0.485\linewidth}
    \centering
    \includegraphics[width=\linewidth]{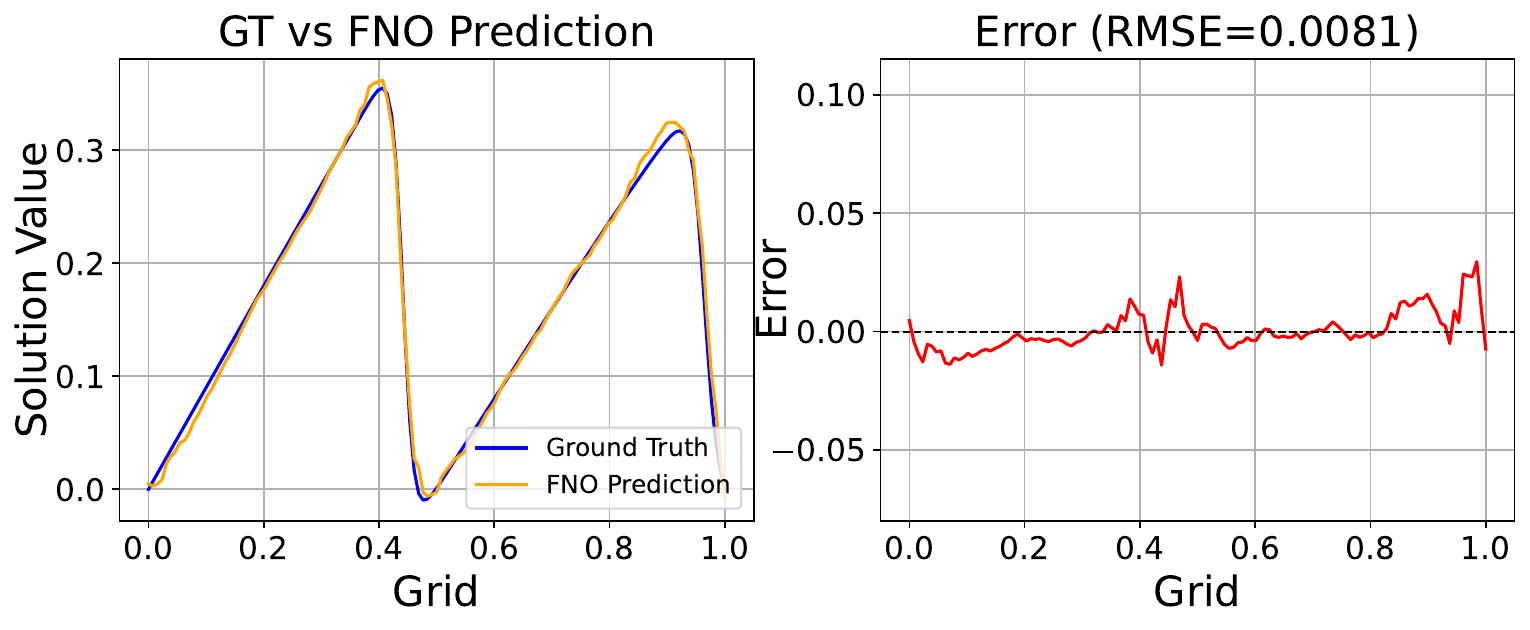}
    \caption{100 sample}
    \label{fig:pred-it2}
  \end{subfigure}

  \vspace{0.6em}

  \begin{subfigure}[t]{0.485\linewidth}
    \centering
    \includegraphics[width=\linewidth]{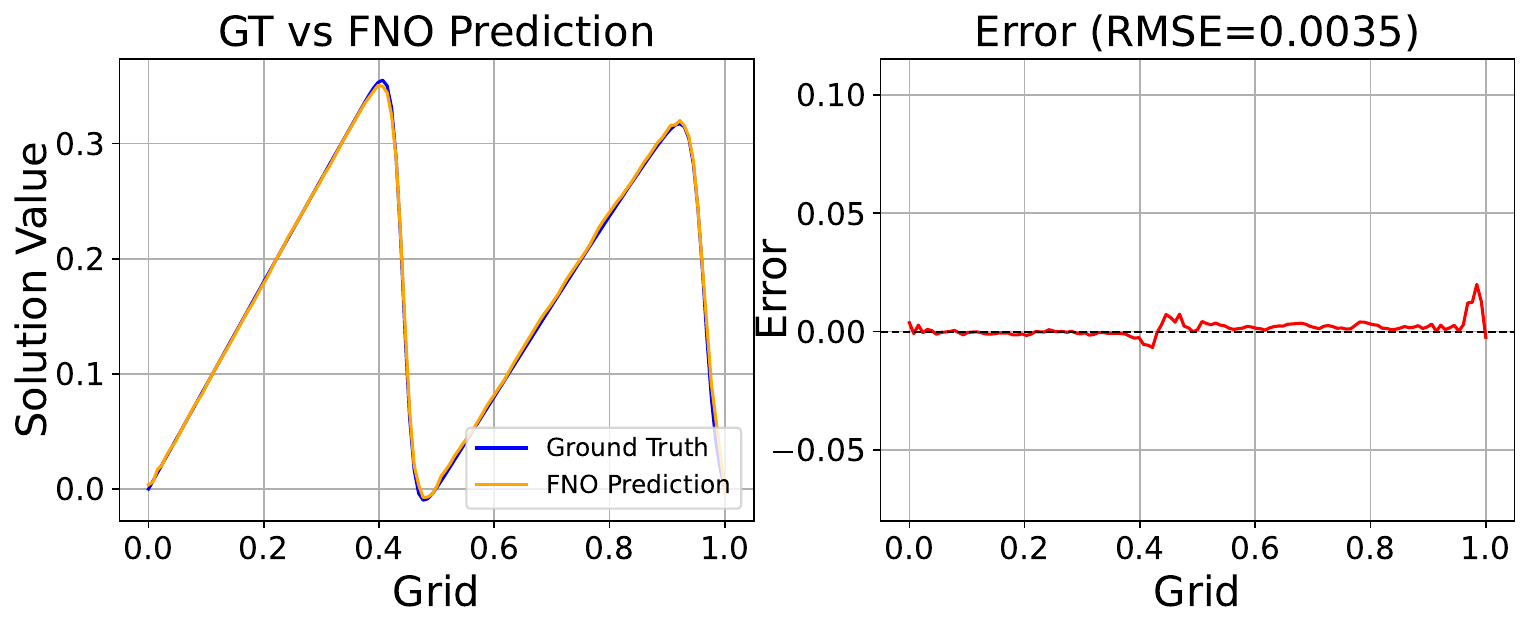}
    \caption{300 sample}
    \label{fig:pred-it3}
  \end{subfigure}\hfill
  \begin{subfigure}[t]{0.485\linewidth}
    \centering
    \includegraphics[width=\linewidth]{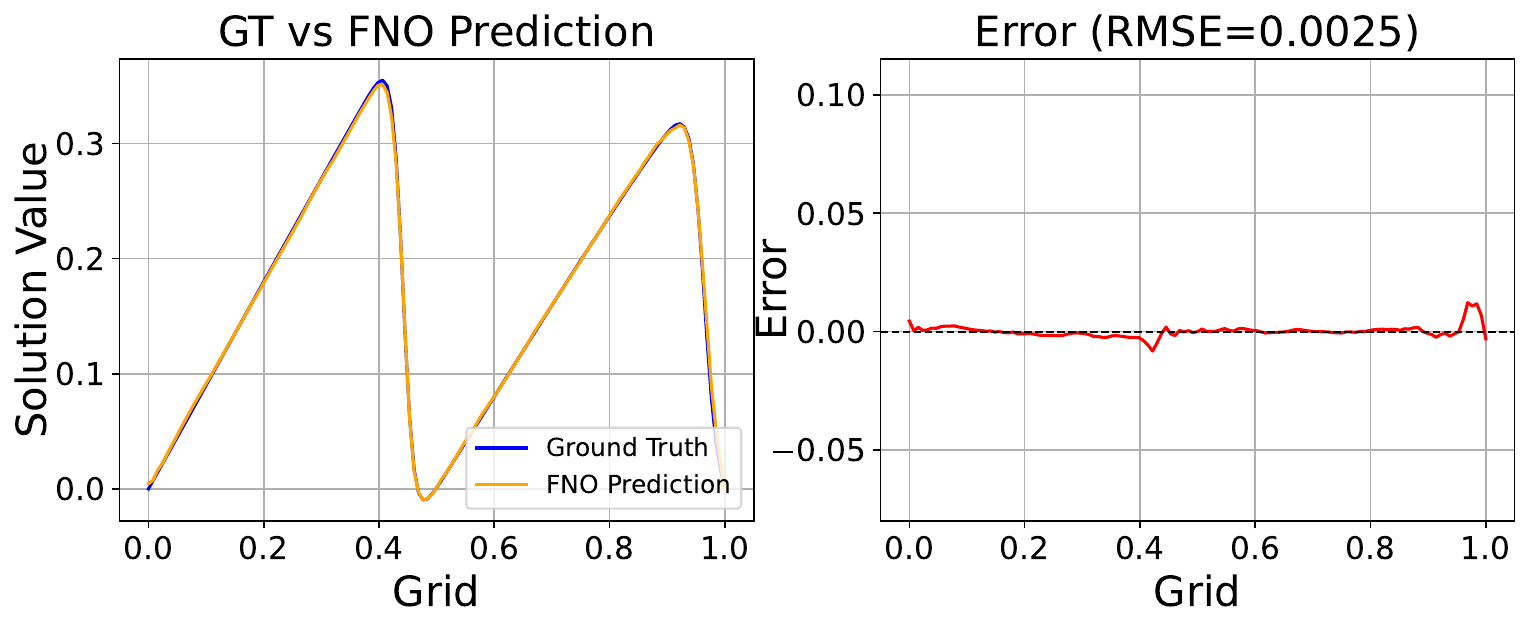}
    \caption{500 sample}
    \label{fig:pred-it4}
  \end{subfigure}

  \vspace{0.6em}

  \begin{subfigure}[t]{0.485\linewidth}
    \centering
    \includegraphics[width=\linewidth]{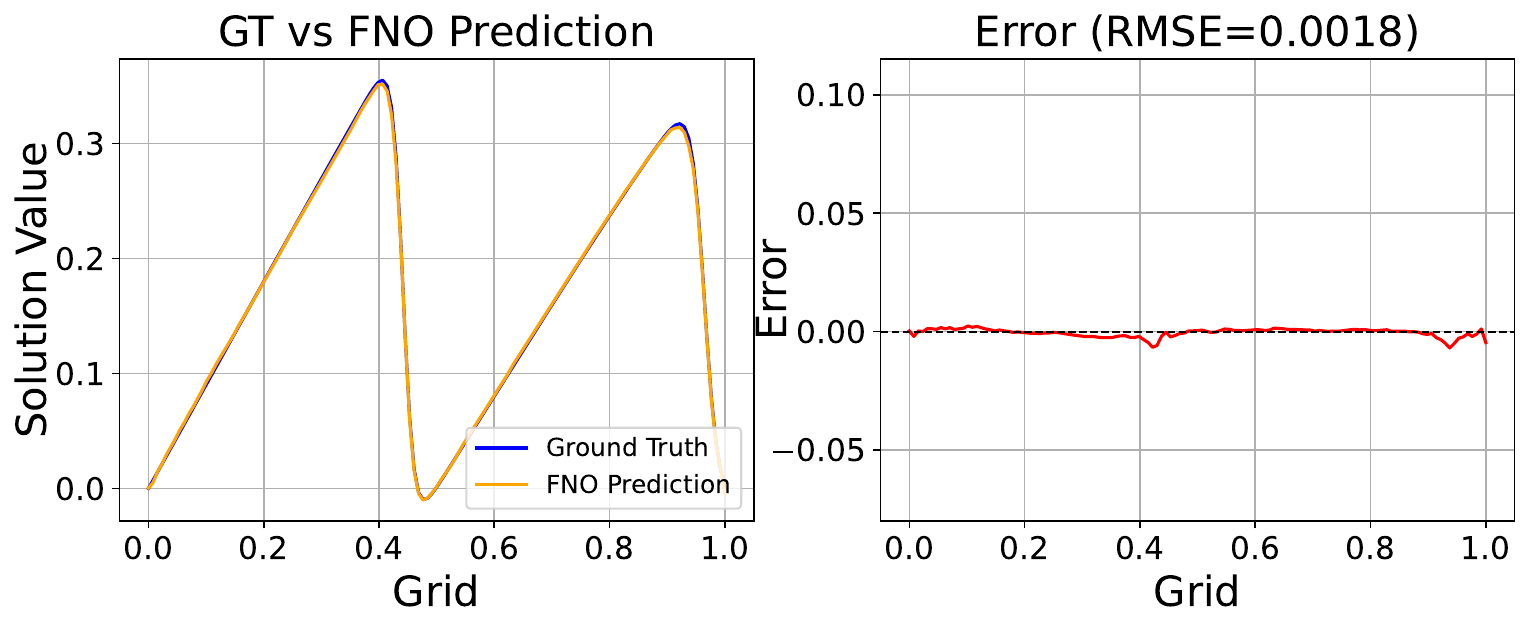}
    \caption{700 sample}
    \label{fig:pred-it5}
  \end{subfigure}\hfill
  \begin{subfigure}[t]{0.485\linewidth}
    \centering
    \includegraphics[width=\linewidth]{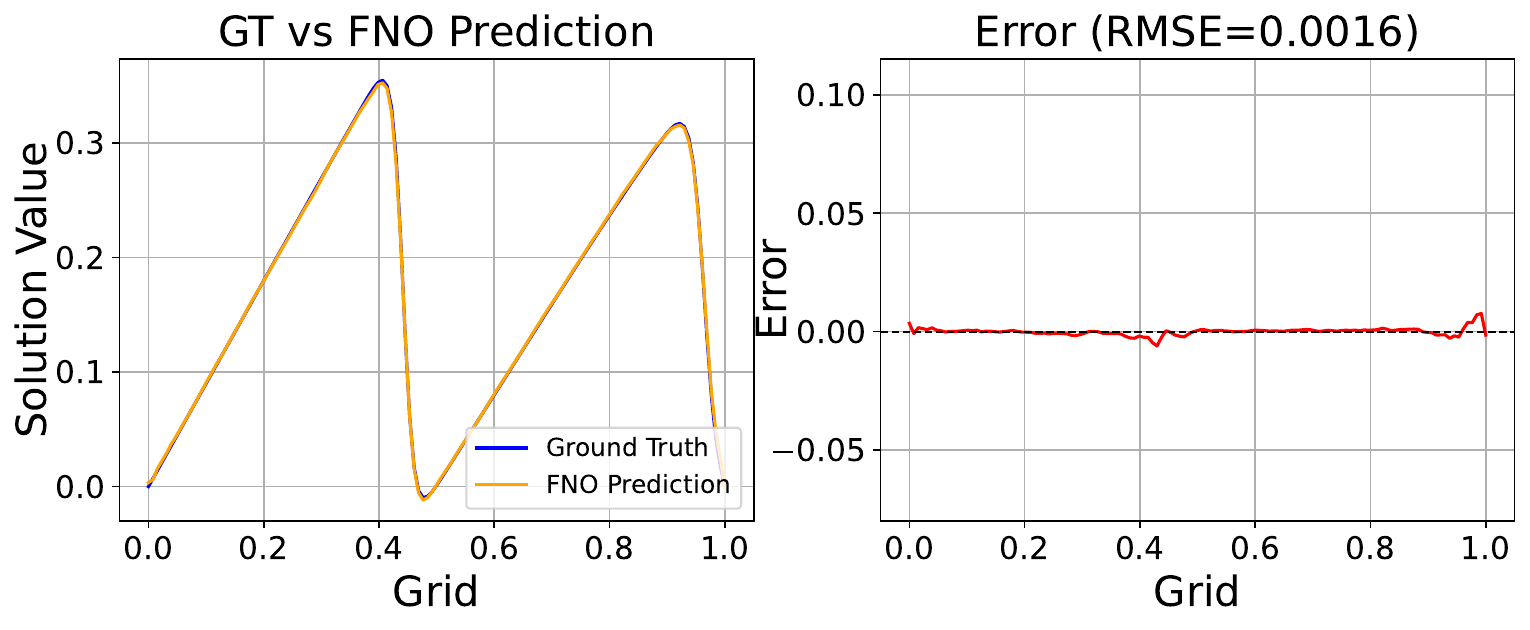}
    \caption{900 sample}
    \label{fig:pred-it6}
  \end{subfigure}

  \caption{Active learning progression: each subfigure shows \emph{(left)} ground truth and model prediction, and \emph{(right)} error map as iterations proceed.}
  \label{fig:al-progression}
\end{figure}

\figref{fig:al-progression} tracks the active-learning trajectory on Burgers using a fixed test instance: each subfigure overlays ground truth and the model prediction (left) and plots the pointwise error (right) with the RMSE noted in the title. Starting from the pretrained model the error is relatively large and oscillatory (RMSE $\approx$ 0.027), but after selecting only 100 samples the mismatch drops substantially ($\approx$ 0.008). Subsequent rounds continue to refine the solution—about 300 samples ($\approx$ 0.0035), 500 samples ($\approx$ 0.0025), 700 samples ($\approx$ 0.0018), and 900 samples ($\approx$ 0.0016)—with the error curve flattening toward zero across most of the domain and remaining concentrated near steep transitions. The sequence illustrates fast early gains from informative sampling followed by diminishing returns, while consistently preserving the overall shape of the solution.

\begin{figure*}[t]
\centering

\begin{subfigure}[t]{0.5\linewidth}
  \centering
  \includegraphics[width=\linewidth]{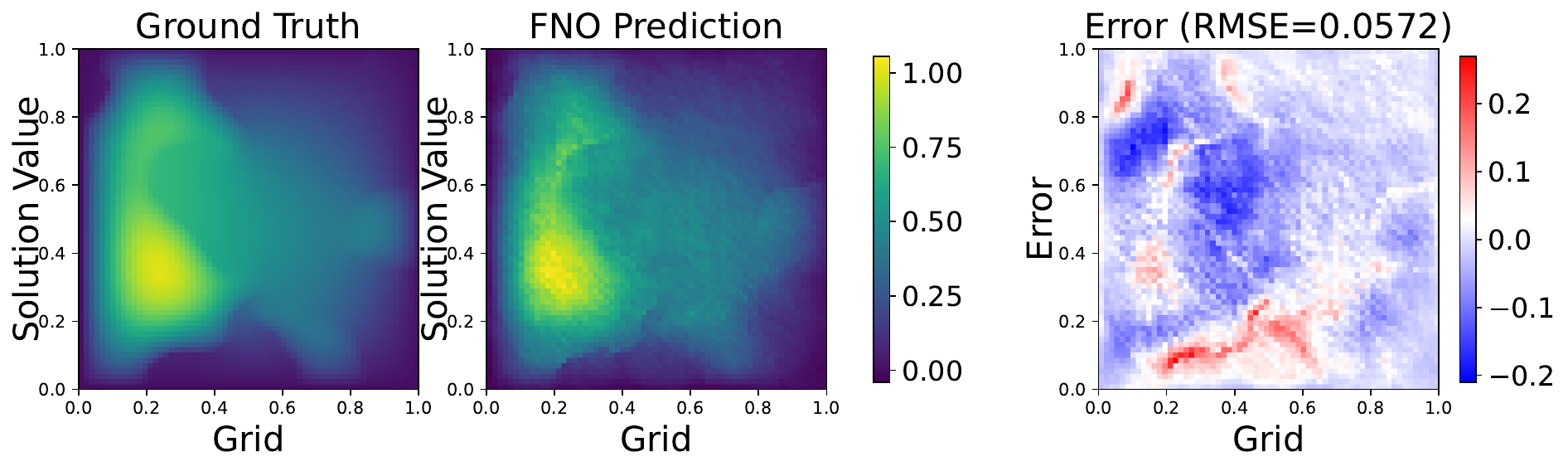}
  \caption{pretrained}
  \label{fig:pred-it1}
\end{subfigure}\hfill
\begin{subfigure}[t]{0.5\linewidth}
  \centering
  \includegraphics[width=\linewidth]{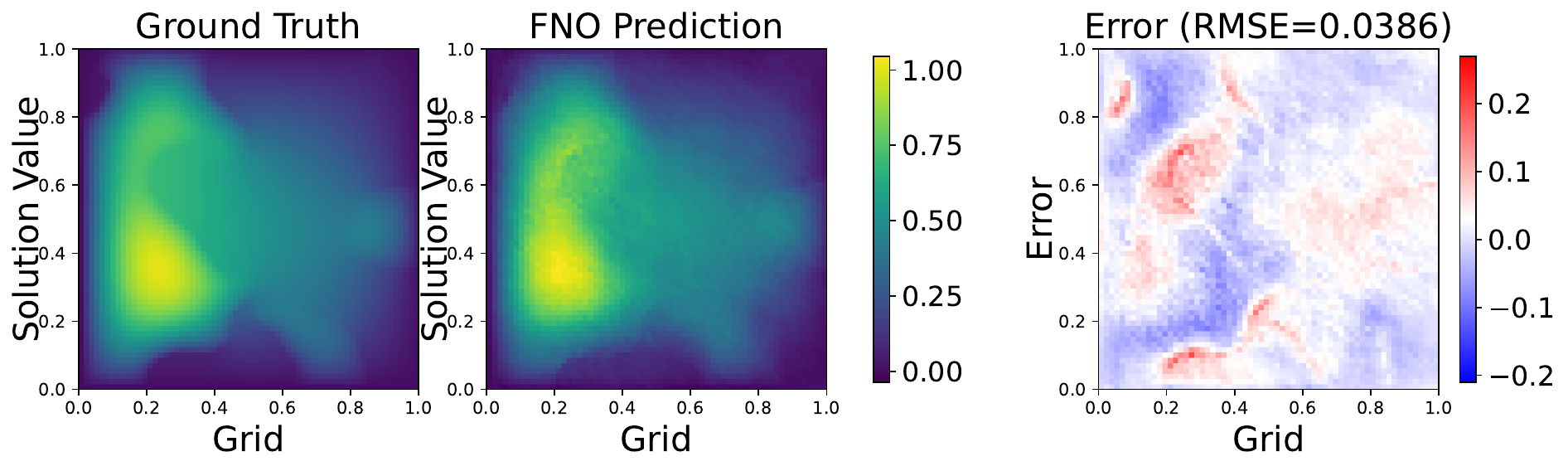}
  \caption{100 samples}
  \label{fig:pred-it2}
\end{subfigure}

\vspace{0.6em}

\begin{subfigure}[t]{0.5\linewidth}
  \centering
  \includegraphics[width=\linewidth]{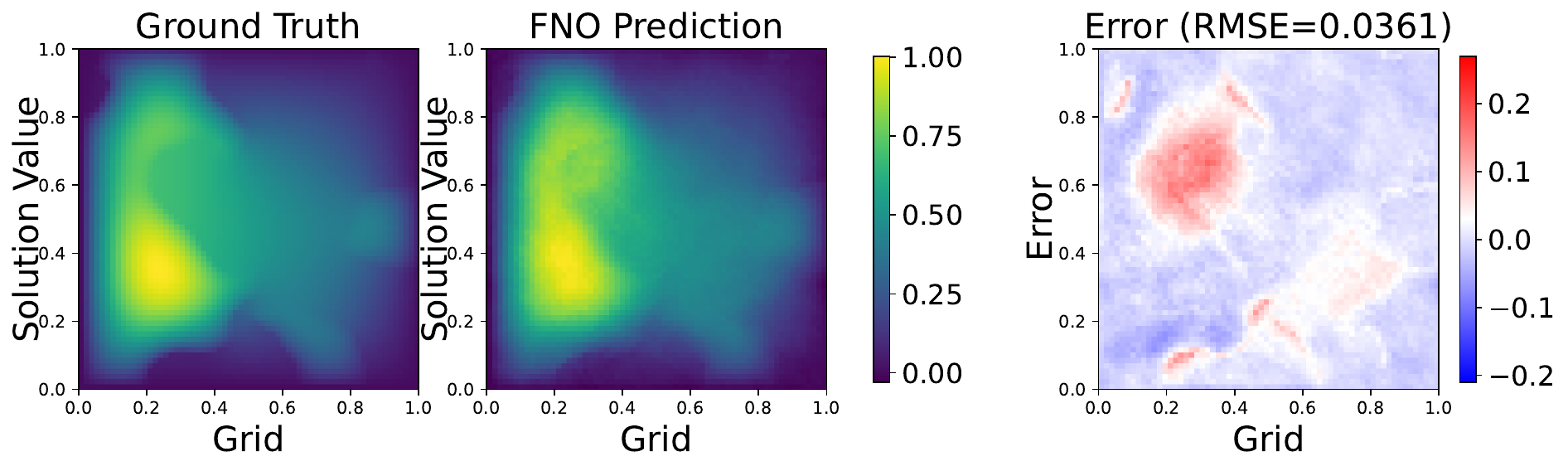}
  \caption{300 samples}
  \label{fig:pred-it3}
\end{subfigure}\hfill
\begin{subfigure}[t]{0.5\linewidth}
  \centering
  \includegraphics[width=\linewidth]{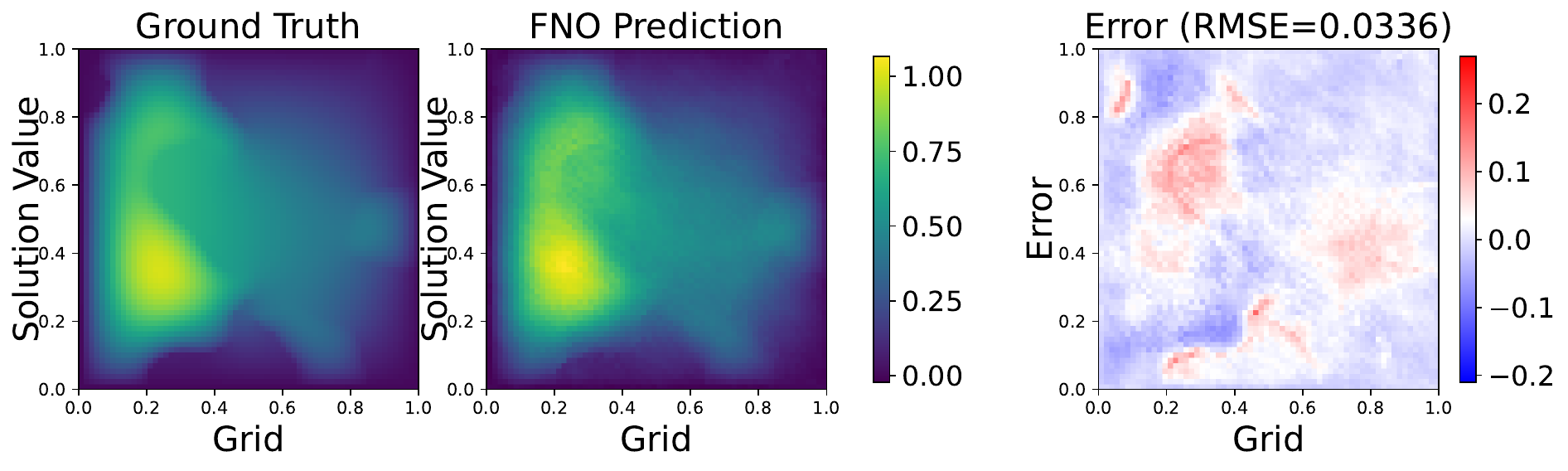}
  \caption{500 samples}
  \label{fig:pred-it4}
\end{subfigure}

\vspace{0.6em}

\begin{subfigure}[t]{0.5\linewidth}
  \centering
  \includegraphics[width=\linewidth]{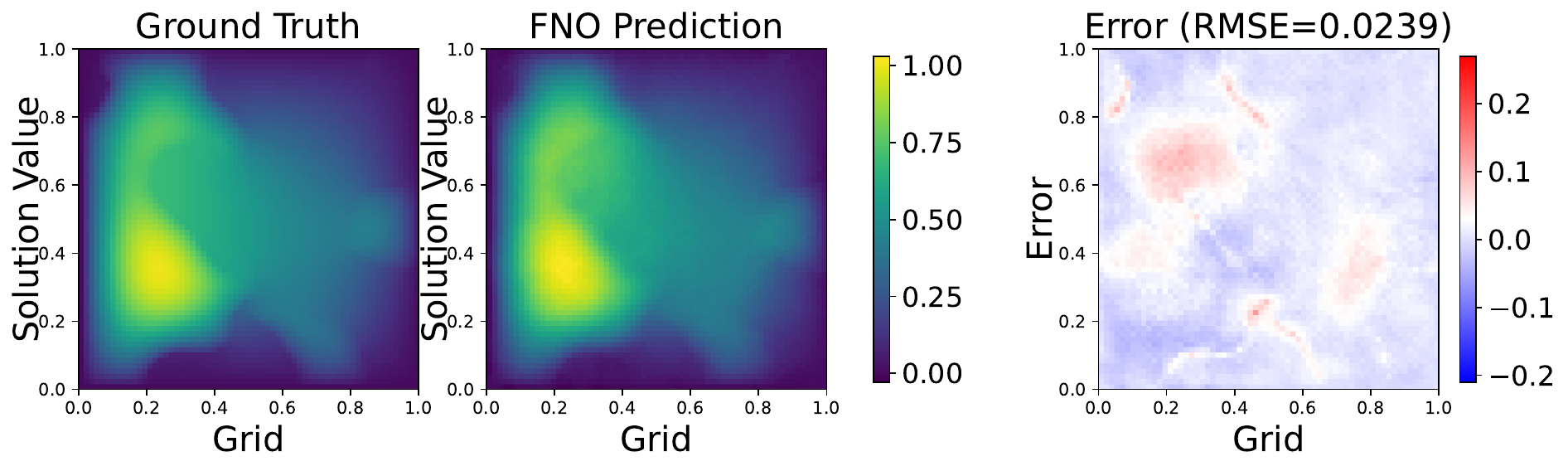}
  \caption{700 samples}
  \label{fig:pred-it5}
\end{subfigure}\hfill
\begin{subfigure}[t]{0.5\linewidth}
  \centering
  \includegraphics[width=\linewidth]{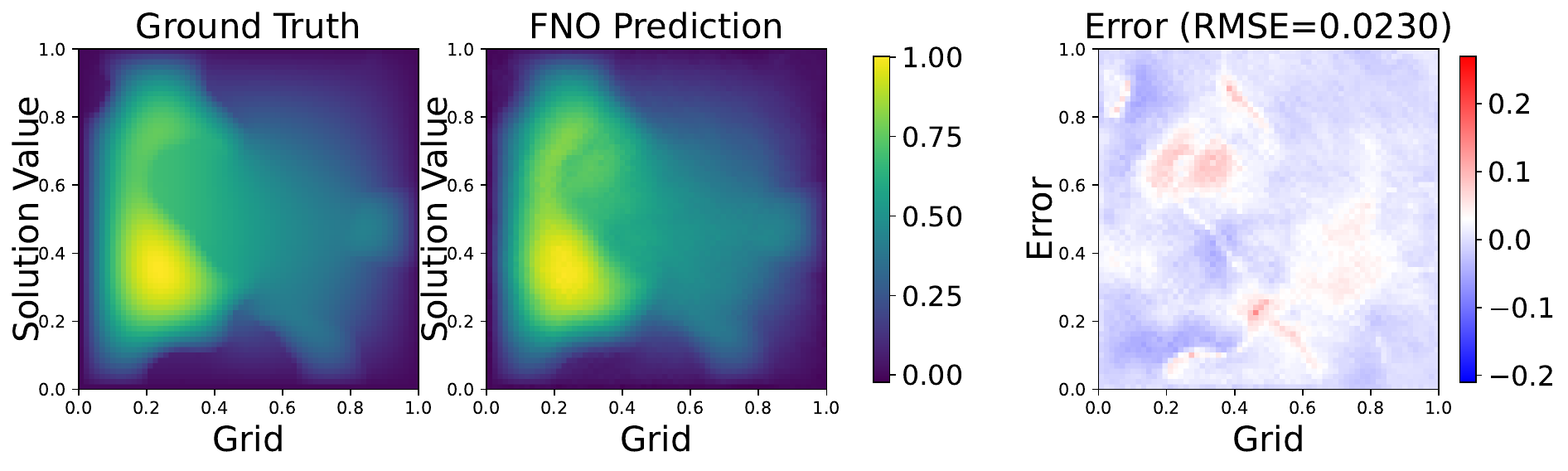}
  \caption{900 samples}
  \label{fig:pred-it6}
\end{subfigure}

\caption{Active learning progression: each subfigure shows \emph{(left)} ground truth, \emph{(middle)} model prediction, and \emph{(right)} error map as the active learning iterations proceed.}
\label{fig:al-progression-darcy}
\end{figure*}

\figref{fig:al-progression-darcy} illustrates active learning on a Darcy test case over six training stages: pretrained, 100, 300, 500, 700, and 900 samples. Each subfigure contains three panels—ground truth on the left, model prediction in the middle (both on a shared color scale), and a signed error map on the right using a symmetric blue-white-red scale. As sampling increases, the prediction progressively matches the ground truth, and the error maps fade from broad, structured errors to mostly low-magnitude residuals. The reported RMSE drops steadily from about 0.057 at initialization to about 0.023 by 900 samples, with the remaining errors concentrated near sharp features and boundary-aligned regions. Overall, the sequence shows rapid early gains from informative data and smaller, consistent improvements at later stages.

\section{PDE Benchmark Formulations}

We provide the precise mathematical formulations of the three benchmark systems used in our experiments: 1D Burgers, 2D Darcy flow, and the Lorenz--96 dynamical system.

\subsection{Burgers' Equation}

We consider the one-dimensional viscous Burgers equation
\begin{equation}
u_t + u u_x = \nu u_{xx}, \quad (x,t) \in [0,1]\times[0,1],
\end{equation}
with periodic boundary conditions and viscosity $\nu=0.002$. 
The learning objective is to approximate the solution operator
\[
u_0(x) \longmapsto u(x,1),
\]
i.e., predicting the terminal-time solution from the initial condition. 
The initial condition is parameterized as
\begin{equation}
u_0(x) = a e^{-a x}\sin(2\pi x)\cos(b\pi x),
\end{equation}
where $a,b \sim \mathcal{U}[1,6]$ are sampled independently.

\subsection{Darcy Flow}

We consider the two-dimensional steady-state Darcy flow equation
\begin{equation}
-\nabla \cdot (c(\mathbf{x}) \nabla u(\mathbf{x})) = f(\mathbf{x}),
\quad \mathbf{x} \in [0,1]^2,
\end{equation}
with homogeneous Dirichlet boundary condition $u(\mathbf{x})=0$ on $\partial[0,1]^2$. 
The forcing term is fixed to $f(\mathbf{x})=1$, and the diffusion coefficient $c(\mathbf{x})>0$ is treated as the input function. 
The learning objective is to approximate the mapping
\[
c(\mathbf{x}) \longmapsto u(\mathbf{x}),
\]
i.e., from coefficient field to steady-state solution. 
Following \cite{Li2020FNO}, coefficient fields are generated by sampling a Gaussian random field and thresholding to obtain a binary permeability field taking values in $\{4,12\}$.

\subsection{Lorenz--96 System}

We consider the Lorenz--96 dynamical system on a one-dimensional periodic lattice of size $N=60$:
\begin{equation}
\frac{d x_i}{dt}
=
(x_{i+1}-x_{i-2})x_{i-1}-x_i+F,
\quad i=1,\dots,N,
\end{equation}
with periodic boundary conditions $x_{i+N}=x_i$. 
The forcing parameter is fixed to $F=4.0$. 
The system is integrated over time horizon $T=1.0$ with time step $\Delta t=0.01$. 
For each instance, we retain only the initial and terminal states $\mathbf{x}(0)$ and $\mathbf{x}(1)$, and the learning objective is to approximate the mapping
\[
\mathbf{x}(0) \longmapsto \mathbf{x}(1).
\]

\end{document}

%% file: macros.tex
\newcommand{\ic}{u_0(x)} 
\newcommand{\instance}{s} 
\newcommand{\trueopm}{\tilde{f}}   
\newcommand{\surrop}{\hat{f}(\cdot; \theta)} 
\newcommand{\proxy}{\phi_\gamma}  

\newcommand{\policy}{\pi}         
\newcommand{\bx}{\boldsymbol{x}} 
\newcommand{\ba}{\boldsymbol{a}} 
\newcommand{\sensorset}{\ba} 
\newcommand{\reward}{r}           
\newcommand{\buffer}{\mathcal{B}} 
\newcommand{\transition}{(s^{(k)},a^{(k)},r,s^{(k+1)})} 

\newcommand{\budget}{B}

\newcommand{\figref}[1]{Fig.~\ref{#1}}

\newcommand{\secref}[1]{\S\ref{#1}}
\newcommand{\appref}[1]{Appendix~\ref{#1}}

%% file: sections/introduction.tex
\section{Introduction}
Deep learning–based surrogates for parametric partial differential equations (PDEs) have emerged as powerful tools for accelerating scientific simulation~\citep{brandstetter2022message, kovachki2023neural, lippe2023pde, yu2024learning, hu2024wavelet, lai2024machine, tang2024learning, jiang2025hierarchical}. 
Popular approaches---such as Fourier neural operators (FNOs)~\citep{Li2020FNO,Li2022GeoFNO}, DeepONets~\citep{lu2021learning}, and more recent graph- and transformer-based operators~\citep{li2020multipole,guibas2021efficient}---learn mappings from PDE inputs (e.g., coefficients, boundary conditions, or initial conditions) to solutions, and have achieved state-of-the-art accuracy across diverse physical systems~\citep{kovachki2023neural}. However, they are inherently data-hungry: training typically requires thousands of high-fidelity numerical simulations on fine spatial grids, each of which carries substantial computational cost. Such reliance on expensive solver calls become a critical bottleneck for practical deployment.


A central inefficiency is that most surrogate training pipelines rely on expensive numerical simulations obtained on dense, structured meshes even when the underlying solutions are strongly inhomogeneous (e.g., shocks in Burgers' equation~\citep{burgers1948mathematical}, high-conductivity channels in Darcy flow~\citep{homsy1987viscous}). Some recent works begin to relax operator learning to simulations with irregular, non-uniform meshes---for instance, Geo-FNO~\citep{Li2022GeoFNO} learns a deformation to handle irregular geometries; MAgNet \citep{boussif2022magnet} learns a continuous, mesh-agnostic PDE solver by combining coordinate-based implicit neural representations with graph neural networks---yet these approaches do not decide where to spend solver queries during training, and still rely on many high-fidelity samples. We therefore ask:
    Can we design a learning algorithm that adaptively allocates solver calls to a small set of mesh grid points per instance, so that we can train a surrogate PDE solver more efficiently, without sacrificing accuracy? 

Designing such a system raises several challenges: 
(i) \emph{Sparse and expensive feedback:} 
Naively measuring surrogate improvement after each newly queried point would entail frequent full retraining, which is computationally prohibitive. (ii) \emph{Budgeted exploration:} 
Under limited budget for solver calls, the adaptive mesh allocation policy must balance exploring new regions with exploiting known informative structures. (iii) \emph{Per-instance adaptivity:} the most informative mesh grid points depend on each instance’s input field and local geometry, so a one-size-fits-all mesh is suboptimal.

\begin{figure*}[t]
    \centering
    \includegraphics[width=\textwidth]{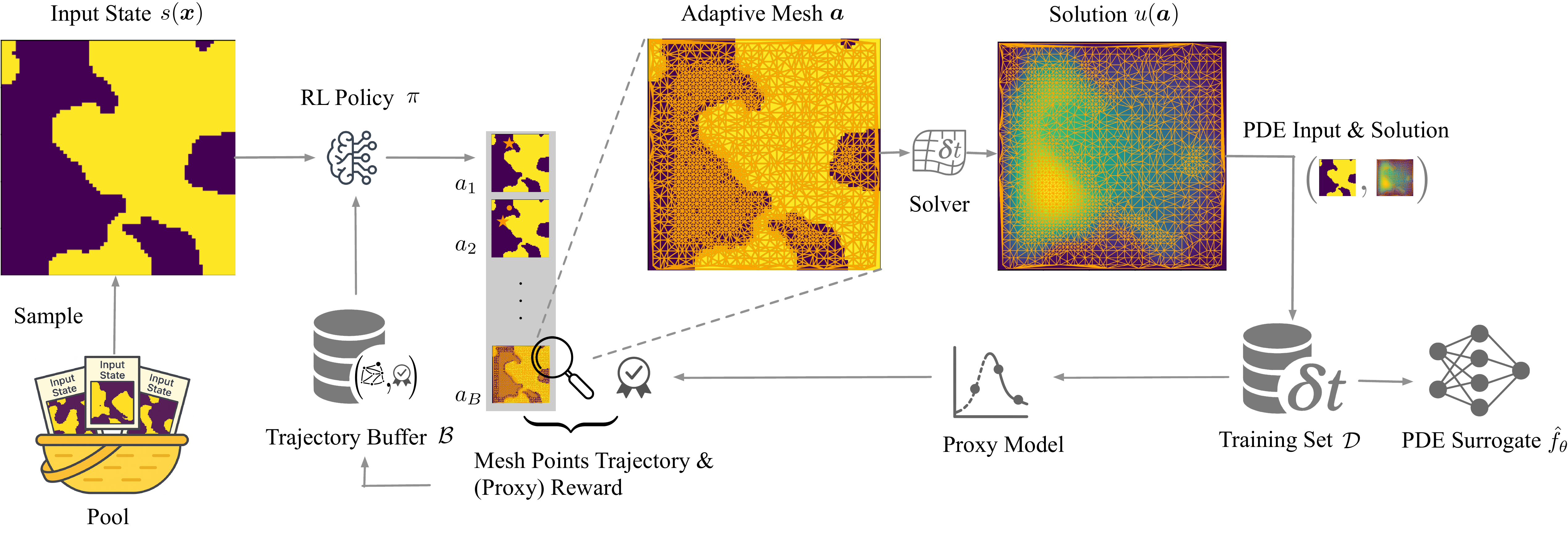}
    \caption{
        The \algname framework. An input state (e.g., PDE initial condition) is sampled from a prior distribution and passed to an RL policy that selects a small set of mesh grid points under a budget. 
        The PDE solver is queried only at these points to produce high-fidelity but sparse observations, which are stored in an updated dataset for surrogate training. 
        A lightweight proxy model provides fast estimates of surrogate improvement and provides reward signal for the RL agent, updating the policy and closing the loop.
    }
    \label{fig:pipeline}
    \vspace{-0.1in}
\end{figure*}

\paragraph{Our approach.}
We introduce \algname, which formulates per-instance \emph{mesh grid point selection} as a sequential decision process (\figref{fig:pipeline}).
For each sampled PDE input state (e.g., an initial condition), a reinforcement learning (RL) policy adaptively selects a small set of mesh grid points; the solver is queried \emph{only} at these locations, yielding sparse but informative observations on a non-uniform mesh that are added to the training set. To avoid prohibitively expensive full retraining as feedback, we introduce a lightweight \emph{proxy model} that predicts the surrogate's improvement attributable to the newly acquired data and supplies a dense, well-aligned reward for RL. 
The surrogate (instantiated as an FNO) is retrained periodically on the accumulated adaptively sampled data, using a simple set-to-grid interface to handle irregular inputs. 
In effect, \algname alternates between two steps: 
(i) \emph{policy improvement} via proxy-rewarded trajectories on new instances, and 
(ii) \emph{surrogate improvement} via periodic retraining on the extended, non-uniform dataset.


Our empirical results on several PDE benchmarks demonstrate that \algname consistently outperforms uniform, random, and several heuristic-based mesh allocation baselines under equal query budgets, achieving significantly lower reconstruction error at substantially reduced simulation cost. With only a fraction of full-grid queries, our surrogate approaches the accuracy of models trained on dense data. 
The proxy model (e.g., kernel ridge regression) exhibits strong correlation with actual surrogate improvements, enabling stable RL training without excessive retraining of the (more computationally expensive) PDE surrogate. Ablations further confirm the benefits of per-instance adaptivity and the importance of proxy alignment.


In summary, our contributions are: 
\begin{itemize}[leftmargin=*]
    \item[\ding{68}] We formulate adaptive, budget-constrained mesh-point selection for PDE surrogates as a sequential decision problem (\secref{sec:pre}).
    \item[\ding{68}] We introduce an end-to-end RL-guided active surrogate learning framework for this problem, where a lightweight proxy model is used to approximate downstream surrogate gains, enabling practical RL reward assignment without frequent full retrains (\secref{sec:method}).
    \item[\ding{68}] We demonstrate empirically that our method achieves superior accuracy–cost trade-offs compared to strong baselines (\secref{sec:experiments}).
\end{itemize}


%% file: sections/related_work.tex
\section{Related Work}\label{sec:related}
\textbf{Active learning for PDE surrogates.}
\looseness -1 A growing body of work has explored active learning (AL) to accelerate PDE surrogate modeling by judiciously selecting training samples. Most existing approaches operate at the \emph{instance} level---i.e. deciding which new PDE instances to simulate next for inclusion in the training set. For example, \citet{Wu2023DMDAL} disentangle instance- and fidelity-level uncertainty via generative modeling, but their acquisition functions are designed for low-dimensional summary outputs and do not natively handle spatially-structured fields. 
More recently, MRA-FNO \citep{Li2023MRAFNO} introduces a cost-aware strategy for selecting both instances and mesh resolutions in training an FNO. By maximizing a utility-cost ratio, MRA-FNO chooses at each step an instance and one of several predefined grid resolutions to simulate, thereby saving data cost. While effective at the dataset level, this approach still confines resolution to a fixed set of global grids and does not perform fine-grained spatial adaptivity \emph{within} each instance.
Similarly, AL4PDE \citep{musekamp2025active} evaluates uncertainty- and feature-based selection strategies (e.g. variance, mutual information), but these acquisitions choose whole instances to add and have no mechanism for adaptive mesh allocation inside an instance's domain.
%
%
In short, existing AL frameworks for neural PDE solvers (e.g., MRA-FNO, AL4PDE) focus on instance-level acquisition—selecting which simulations or global resolutions to run under the assumption that each selected instance yields a full solution field—whereas RLMesh addresses a complementary and previously unmodeled axis of adaptivity: per-instance spatial point selection under a strict local query budget.

\textbf{Adaptive mesh refinement.} 
Adaptive mesh refinement (AMR) is a classical strategy in
numerical PDE solvers \citep{melia2025hardware}: the mesh is locally refined in regions of large error, steep gradients, or discontinuities, while keeping a coarse grid elsewhere to save cost; 
Early AMR methods introduced structured grid refinement with hierarchical data structures such as quadtrees/octrees~\citep{BergerOliger1984,BergerColella1989}, and used error estimators or physics-based heuristics to decide where to refine. 
In the finite element community, \emph{hp}-adaptivity provides an additional lever: solvers reduce element size (\emph{h}) and/or increase polynomial degree (\emph{p}) according to local regularity, yielding exponential convergence on suitably smooth subregions \citep{Demkowicz2006hp}. 
Traditional methods use handcrafted error indicators or solve auxiliary equations to adapt the mesh during a single forward run for improved accuracy.

Recently, there is growing interest in using deep RL to \emph{learn} mesh adaptation policies automatically.  \citet{Foucart2023} formulate AMR as a sequential decision process in which an agent decides whether to refine or coarsen individual grid cells, and show that an RL policy can replicate classical refinement strategies without explicit error estimators. 
\citet{Yang2021RLAMR} extend this to global mesh views with variable-size state/action spaces, using specialized architectures (including graph-based) to scale across meshes.  \citet{Rowbottom2024} study mesh \emph{relocation} ($r$-adaptivity) and use graph neural networks trained via differentiable finite element solvers to move mesh nodes and minimize the PDE solution error directly. \citet{SwarmRL2024} present \emph{SwarmRL}, a modular framework for multi-agent reinforcement learning in noisy physical environments, further illustrating the broader push toward RL-driven adaptivity in scientific systems. All of these methods tackle adaptivity for improving a solver's result \emph{during a single simulation}. 
In contrast, our work uses adaptivity to improve \emph{learning efficiency across many instances}: Instead of refining a mesh to solve one PDE instance accurately, we adaptively sample each instance's solution to collect the most informative training data for a surrogate model. 
This shifts the focus from solver-time to training-time efficiency and introduces a new feedback mechanism, using a proxy model to estimate surrogate improvement instead of relying on PDE residuals.

%% file: sections/problem_formulation.tex
\section{Problem Setup}
\label{sec:pre}
\paragraph{Neural operator.} Let $\Omega \subset \mathbb{R}^{d}$ 
denote the spatial coordinates (e.g., $d=1$ represents the Burgers' equation; $d=2$ for 2D Darcy flow) and $t \in [0,T]$ denote the physical time. 
We consider the PDE in its generic form written as:
\begin{align}
    G\big(u, \partial_x u, \partial_t u, \dots, x, t, c(x)\big) = 0 \\
    \quad x \in \Omega \nonumber
\end{align}
subject to the governing equation $G$, initial condition $\ic$, and spatially varying coefficients $c(x)$.
For a time-discretized system, we view the solution at each time step $t$ as a spatial field $u(x)$.
The PDE solver then induces a solution operator $f$ that, given the input functions (e.g., initial conditions, coefficients), returns the full solution:
\begin{equation}
    f: \mathcal{S}\to\mathcal{U}, \qquad s(\cdot)\mapsto u(\cdot)
\end{equation}
where $s \in \mathcal{S}$ is the input function $s: \Omega \to \mathbb{R}^{d_s}$ (e.g., $\ic$, $c(x)$),
and $u \in \mathcal{U}$ is the output function $u: \Omega \to \mathbb{R}^{d_u}$.
In practice, $\Omega$ is discretized by $n$ points $\Omega_n \equiv \boldsymbol{x} := \{x_1, x_2, \dots, x_n\}$ with $x_i \in \mathbb{R}^d$. 
Then the functions are represented as finite-dimensional tensors $s(\boldsymbol{x}) \in \mathbb{R}^{n \times d_s}$ and $u(\boldsymbol{x}) \in \mathbb{R}^{n \times d_u}$.

The neural operator aims to train a network $\hat{f}(\cdot; \theta)$, with parameters $\theta$, to approximate the solver $f$ \citep{Li2020FNO, wen2022u}.

\paragraph{RL-guided mesh optimization.}  Our goal is to learn an adaptive policy for mesh point selection 
$\policy : \mathcal{S} \times 2^{\Omega_n} \to \Omega_{n}$ for the PDE surrogate training.
The input of $\policy$ includes the input function to the neural operator and the subset of mesh grid points already chosen; the output of $\policy$ corresponds to an action
$a \in \Omega_n$  as the next mesh grid point in spatial domain $\Omega_n$ to be selected.

The objective is to learn $\hat{\policy}$ that minimizes expected surrogate error:
\begin{equation}
    \hat{\policy} = \arg\min_\policy \mathbb{E}\big[\ell(\hat{f}(s(\bx); \theta_\policy), f(s(\bx)))\big].
\end{equation}

We assume there exists a numerical solver that can be queried online (see \figref{fig:pipeline}). In addition, we define a per-instance budget 
$B$ as the maximum number of mesh points allowed for each PDE instance. Querying the solver at chosen mesh points $\ba := \{a_j\}_{j=1}^B$, we update the PDE surrogate model $\hat{f}(; \theta)$ using the training dataset containing historical samples and the newly arrived data pair $(s(\bx), u(\ba))$. Here $u(\ba):= \tilde{f}(s(\bx), \ba)$, with a solver $\tilde{f}$ taking non-structured meshes to focus on the solution at the mesh grid points $\ba$.

The model is evaluated using RMSE loss on high-resolution validation instances
\begin{align}
\ell(\hat{f}\big(s(\bx); \theta), f(s(\bx)\big) = 
\sqrt{\frac{1}{M}\sum_{j=1}^{M} 
\big(\hat{f}(s^{(j)}; \theta) - f(s^{(j)}\big))^2 }. \nonumber
\end{align}
where $s^{(j)}:= s^{(j)}(\bx)$ represents the input state at $j$-th sample, and $M$ denotes the size of the validation dataset.


%% file: sections/method.tex
\section{RL-Guided Mesh Optimization}
\label{sec:method}
In this section, we present \algname, an RL-guided active acquisition framework that learns. For each PDE instance, we select a small number of meaningful mesh points so that a downstream neural operator surrogate can be trained efficiently under a fixed budget.

\subsection{High-Level Workflow}
Following \secref{sec:pre}, let $s(\bx)$ denote the PDE input state and $\trueopm$ the numerical solver that returns the field $u(\cdot)$.
For each sampled instance $s$, our goal is to sequentially choose a set of mesh points $\ba \subseteq \Omega_{n}$ under a per-instance budget $|\ba| \leq B$ and query the solver only at those locations; the resulting sparse measurements are added to a growing training set for the surrogate $\surrop$.
To avoid full surrogate retraining after each mesh grid selection, we fit a proxy model that estimates the improvement of $\surrop$ due to the newly acquired data and uses that estimate as the reward for the RL policy.
Periodically, we retrain $\surrop$ on all data collected so far.

Algorithm~\ref{alg:overall_protocol} shows the high-level workflow (as was illustrated in \figref{fig:pipeline}): \algname operates as an iterative loop where a new PDE instance is sampled, a set of mesh points is selected according to the current policy, the solver is queried at these locations to obtain sparse observations, a proxy reward is computed to guide policy optimization, and the surrogate is periodically retrained on the accumulated non-uniform data.

\begin{algorithm}[t]
\caption{Learning Protocol}
\label{alg:overall_protocol}
\KwIn{
  Seed dataset $\mathcal{D}_0$; 
  RL trajectory buffer storing transitions $\buffer:=(s,\ba,r,s') $ ($s$ shows the current state, $\ba$ is the action denoting the selected mesh grid points, $r$ is the reward, $s'$ is the next state); 
  PDE solver $\trueopm$; 
  Surrogate model $\surrop$ with initial parameters $\theta^{(0)}$; 
  Initial RL policy $\policy^{(0)}$; 
  Iteration number $K$; Per-instance Budget $\budget$
}
Initialize: $\mathcal{D} \gets \mathcal{D}^{(0)}$, $\budget$, $\buffer \gets \emptyset$\;
\For{$k = 1, \dots, K$}{
    Randomly sample an input state $\instance^{(k)}$\;
    
    Select mesh points $\sensorset^{(k)}$ using current policy $\policy^{(k-1)}$\ under budget $B$; 

    Query PDE solver: $  (\instance^{(k)}, \sensorset^{(k)}) \gets \trueopm(\instance^{(k)}, \sensorset^{(k)})$

    Update dataset: $
    \mathcal{D}^{(k)} \gets \mathcal{D}_{k-1} \cup \{(\instance^{(k)},  \sensorset^{(k)})\}$

    Update trajectory buffer: $    \buffer^{(k)} \gets \buffer_{k-1} \cup \transition$

    Update RL policy $\policy^{(k)}$ with $\buffer^{(k)}$
    
    Retrain surrogate model: $
    \surrop \;\gets\; \text{train}(\mathcal{D}^{(k)})$
}
\textbf{Return:} Final dataset $\mathcal{D}^{(K)}$, final surrogate parameters $\theta^{(K)}$
\end{algorithm}

\subsection{RL for Adaptive Mesh Grid Selection}

\paragraph{Formulation:}
We formulate the adaptive mesh grid selection as a finite-horizon Markov Decision Process (MDP) in which an agent sequentially selects cells of the mesh grid under a fixed per-instance budget $B$.
Each episode corresponds to one PDE instance and proceeds for $B$ steps.
At each step $k$, the environment is represented by a state $s^{(k)}$ that concatenates a binary mask indicating which cells have already been selected, and a normalized encoding of the PDE input.

The agent's action space at step $k$ consists of all grid cells not yet selected.
When the policy selects an action $a_k$, the corresponding cell is marked as selected and the state is updated to reflect the new mask and any contextual features derived from the partially chosen grid.
This process continues until the budget $B$ is exhausted, yielding a sequence of grid selections $(a_1, \dots, a_B)$ for the given PDE instance.

\paragraph{Reward construction.}
Rather than providing step-wise feedback, the environment issues a \textit{single terminal reward} once the full set of $B$ mesh grid points has been chosen.
This reward is designed to measure the expected benefit of the selected mesh for training a downstream surrogate solver $\surrop$.
Because retraining $\surrop$ after each episode to obtain an exact reduction would be computationally prohibitive, we introduce a lightweight \textit{proxy model} that predicts the surrogate's generalization error given a proposed mesh.

\begin{figure}
    \centering
    \includegraphics[width=0.9\linewidth]{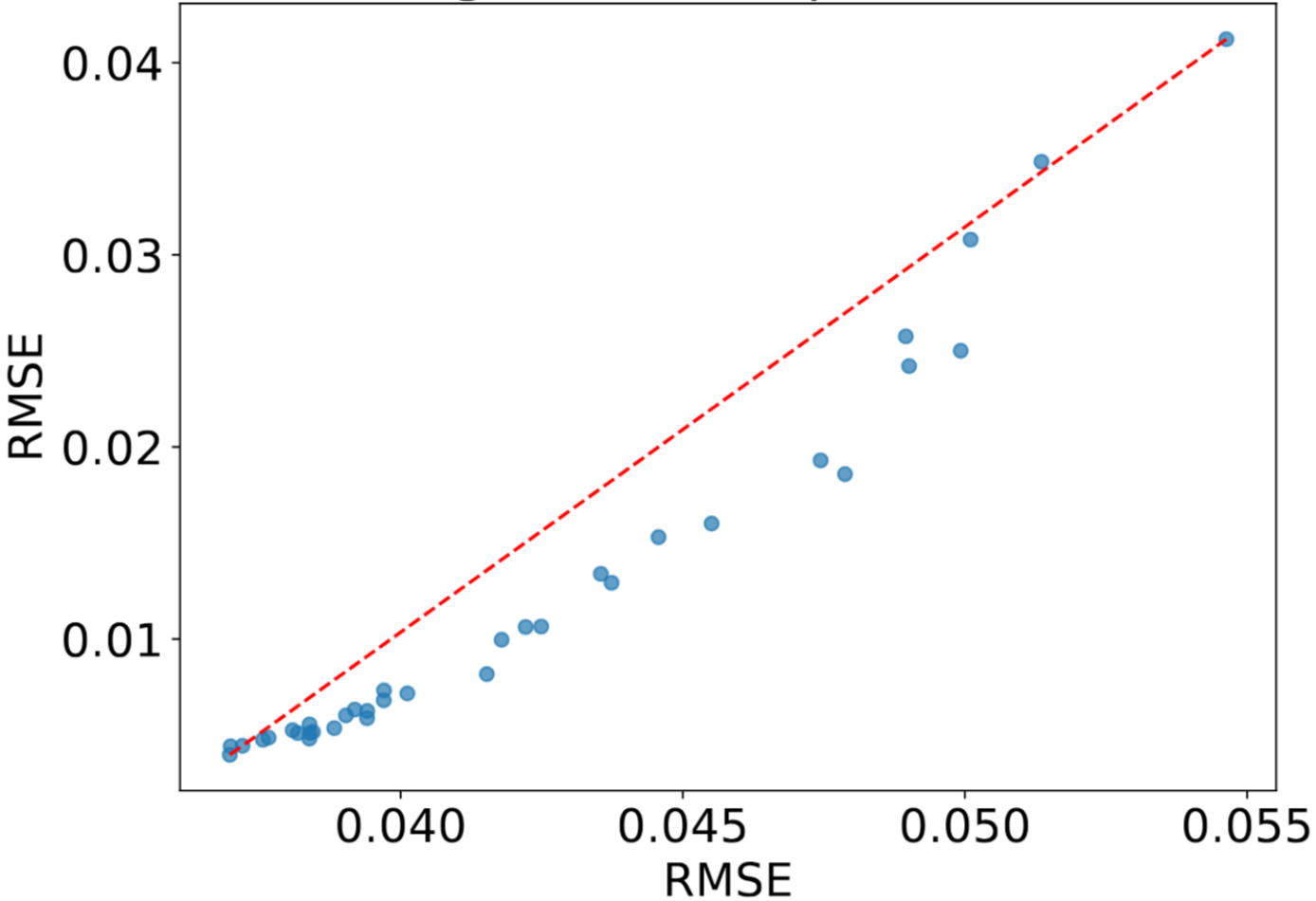}
    \caption{Correlation between proxy (RBF kernel ridge, x-axis) and FNO surrogate errors (y-axis) across subset sizes. Each point corresponds to a retrained model. Spearman correlation (0.9908) indicates strong rank consistency. }
    \label{fig:kernel_ridge}
\end{figure}

In practice, the proxy is implemented as kernel ridge regression with a closed-form solution
\begin{equation}
    \hat{g}(s) = \sum_{j=1}^{N} \alpha_j \, k\big(s(\bx), s^{(j)}(\bx)\big), \quad \boldsymbol{\alpha} = (\boldsymbol{K} + \lambda I)^{-1} \boldsymbol{u},
\end{equation}
where $\boldsymbol{K}$ is the kernel matrix evaluated on the training inputs $s^{(j)}$, and $\boldsymbol{u}$ denotes the vector of known function values.
The proxy model allows fast updates and inference while maintaining a strong correlation with the true surrogate error. It is only used after an initial pretraining phase in which the FNO surrogate is trained on 100 fully observed instances, ensuring a stable warm-start regime for acquisition. In this regime, the RBF kernel ridge proxy exhibits near-perfect monotonic agreement with the FNO surrogate across subset sizes (Spearman $\rho = 0.9908$; see \figref{fig:kernel_ridge}). This high rank correlation indicates that the proxy preserves the relative ordering of mesh selections in terms of downstream surrogate error, ensuring that reward signals remain well aligned with true generalization improvement. As a result, the RL agent can optimize placements using the proxy without frequent expensive retraining of the full surrogate. Additional details on proxy selection and evaluation are provided in \appref{sec:proxy}.

For each episode, we compute the proxy-predicted error of the surrogate before and after adding the instance with $B$ newly selected mesh grid points, denoted as $\epsilon_\text{old}$ and $\epsilon_\text{new}$.
The terminal reward is then defined as
\begin{equation}
    r = -\kappa(\epsilon_\text{new} - \epsilon_\text{old}),
\end{equation}
with a scaling factor $\kappa$. It is then rescaled to the interval $[-1, 1]$ to stabilize the learning.
Concentrating the reward at the end of the episode directly aligns the signal with the goal of improving surrogate accuracy, while avoiding hand-crafted local heuristics such as uncertainty sampling.
Throughout the paper, the surrogate model refers to a Fourier Neural Operator trained to approximate the PDE solution operator, while the proxy model refers to a lightweight kernel ridge regression estimator used solely for reward computation during RL training.

\paragraph{Training.}
The policy is trained using a value-based RL algorithm, instantiated as a deep Q-network with experience replay and a target network under a greedy exploration schedule.
Episodes thus generate trajectories of states and actions with a single terminal reward supplied by the proxy, enabling efficient policy optimization without retraining the surrogate at every step.
To keep the proxy well aligned with the true surrogate error, the surrogate $\surrop$ is periodically retrained on the accumulated non-uniform data, amortizing its cost while ensuring that the proxy remains informative.
The complete \algname is also detailed in Algorithm~\ref{alg:passive_rl_batch}.
\begin{algorithm}[t]
\caption{\algname Algorithm}\label{alg:passive_rl_batch}
\KwIn{Initial FNO $\surrop$, initial proxy model $\hat{g}$, initial RL policy $\policy^{(0)}$, iteration number $K$, per-instance budget $\budget$, retraining interval $L$}

Initialize: $\mathcal{D} \gets \mathcal{D}^{(0)}$,  
trajectory buffer $\buffer \gets \emptyset$.

\For{$k = 1, \dots, K$}{

    Select input states $s^{(k)}$ randomly

    Select $\budget$ mesh grid points \(\sensorset^{(k)}=(a^{(k)}_1,\dots,a^{(k)}_{B})\) using current RL policy $\policy^{(k-1)}$

    Query solver to obtain solution on a non-uniform grid: $u(\ba^{(k)}) \gets \trueopm(s^{(k)}, \ba^{(k)})$

    Update dataset: 
    $\mathcal{D}^{(k)} \gets \mathcal{D}^{(k-1)} \cup (s^{(k)}, u(\sensorset^{(k)}))$ \\
    Compute proxy-based terminal reward $R_\text{batch}$ using $\hat{g}$. 

    Update RL dataset: $\buffer \gets \buffer \cup
    \bigl(s^{(k)}_{b},\,a^{(k)}_{b},\,0,\,s^{(k)}_{b+1} \bigr), \forall b\leq B-1$,\\
    $\buffer \gets \buffer \cup \bigl(s^{(k)}_{B},\,a^{(k)}_{B},\,R_\text{batch}\bigr)$ 

    Update RL policy $\policy^{(k)}$ (e.g. via deep Q-learning) using batch trajectories $\buffer$

    Retrain proxy model using current dataset: $\hat{g} \gets \text{train}(\mathcal{D}^{(k)})$

    \If{$k \mod $L$ = 0$}{
        Retrain FNO model using current dataset: $\surrop \gets \text{train}(\mathcal{D}^{(k)})$
    }
}
\textbf{Return:} Final dataset $\mathcal{D}^{(K)}$, trained RL policy $\policy^{(K)}$, trained FNO $\surrop$.
\end{algorithm}

%% file: sections/experiments.tex
\section{Experiments}
In this section, we show our empirical verification on three PDE families, 1D Burgers (terminal-time prediction), 2D Darcy (steady mapping), and Lorenz-96 system (chaotic lattice system).
\label{sec:experiments}

\subsection{Experimental Setup}

\begin{figure*}[t]
  \centering
  \begin{subfigure}[t]{0.325\linewidth}
    \centering
    \includegraphics[width=\linewidth]{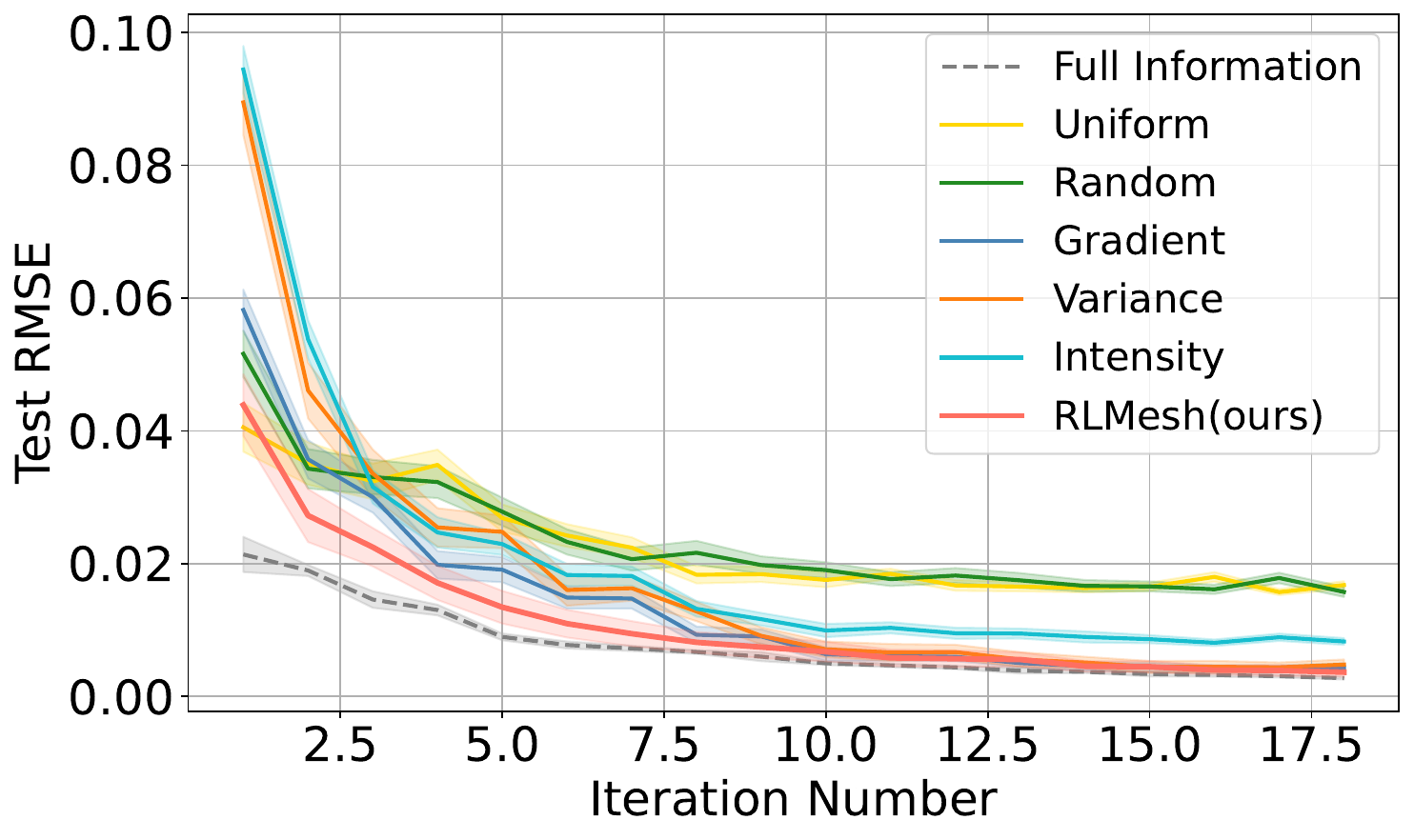}
    \caption{Burgers, RMSE vs. Iterations}
    \label{fig:burgers_main_result}
  \end{subfigure}
  \begin{subfigure}[t]{0.325\linewidth}
    \centering
    \includegraphics[width=\linewidth]{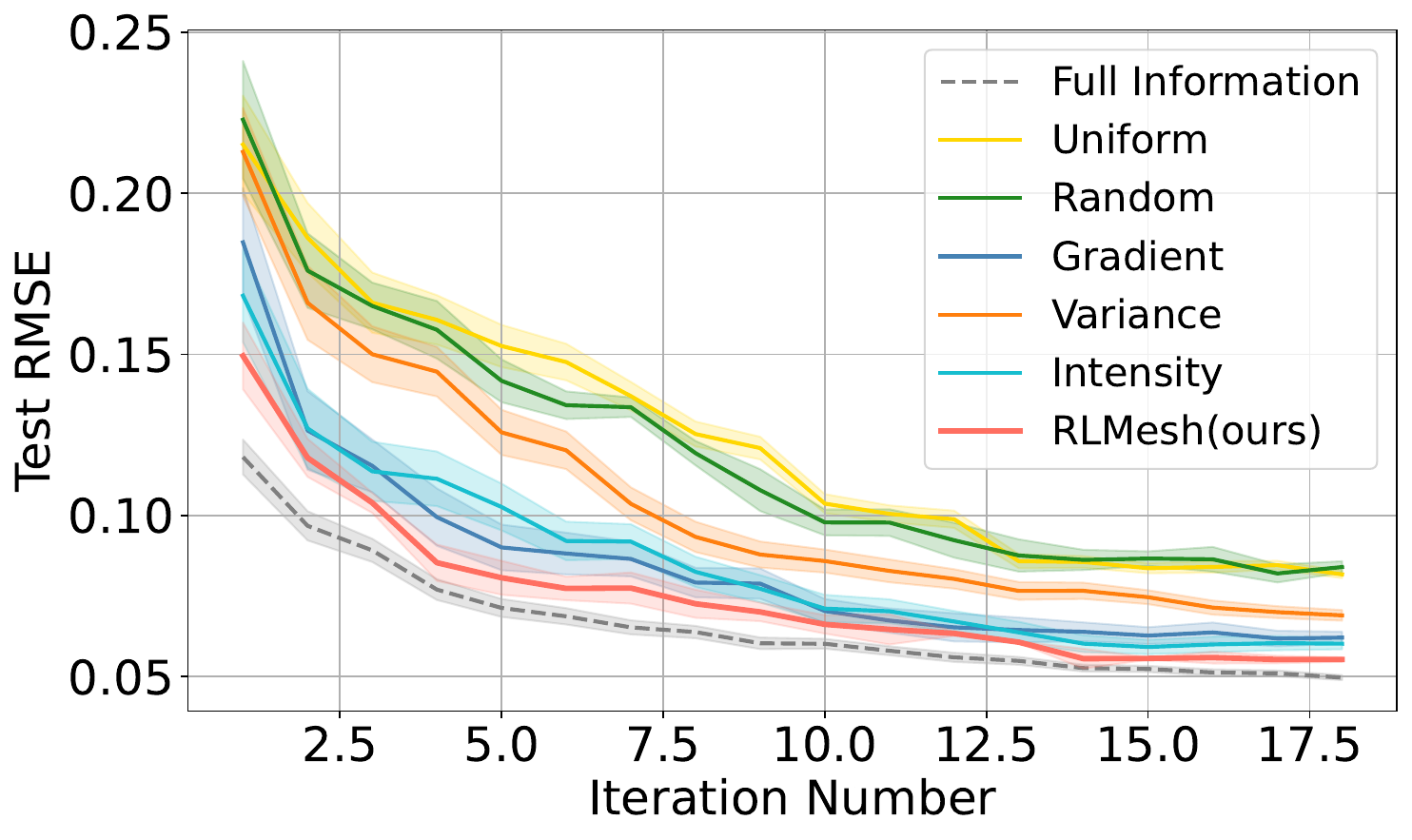}
    \caption{Darcy, RMSE vs. Iterations}
    \label{fig:darcy_main_result}
  \end{subfigure}
  \begin{subfigure}[t]{0.325\linewidth}
    \centering
    \includegraphics[width=\linewidth]{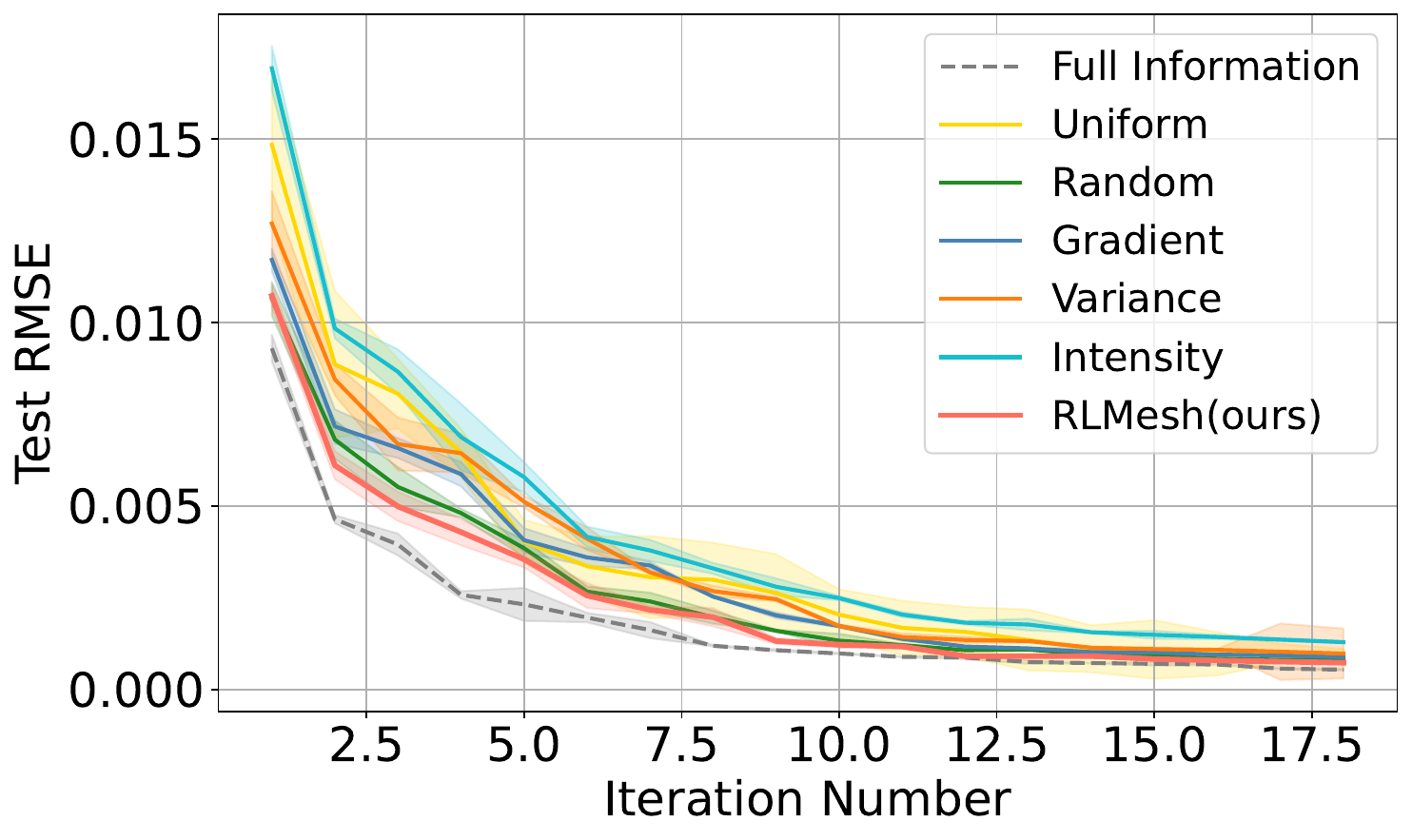}
    \caption{Lorenz--96, RMSE vs. Iterations}
    \label{fig:lorenz_main_result}
  \end{subfigure}
  \vspace{-0.2em}
  \caption{Active learning performance across three benchmark systems. \algname consistently outperforms heuristic baselines under identical query budgets.}
  \label{fig:main_results_all}
\end{figure*}

\begin{figure}[t]
  \centering
  \includegraphics[width=\linewidth]{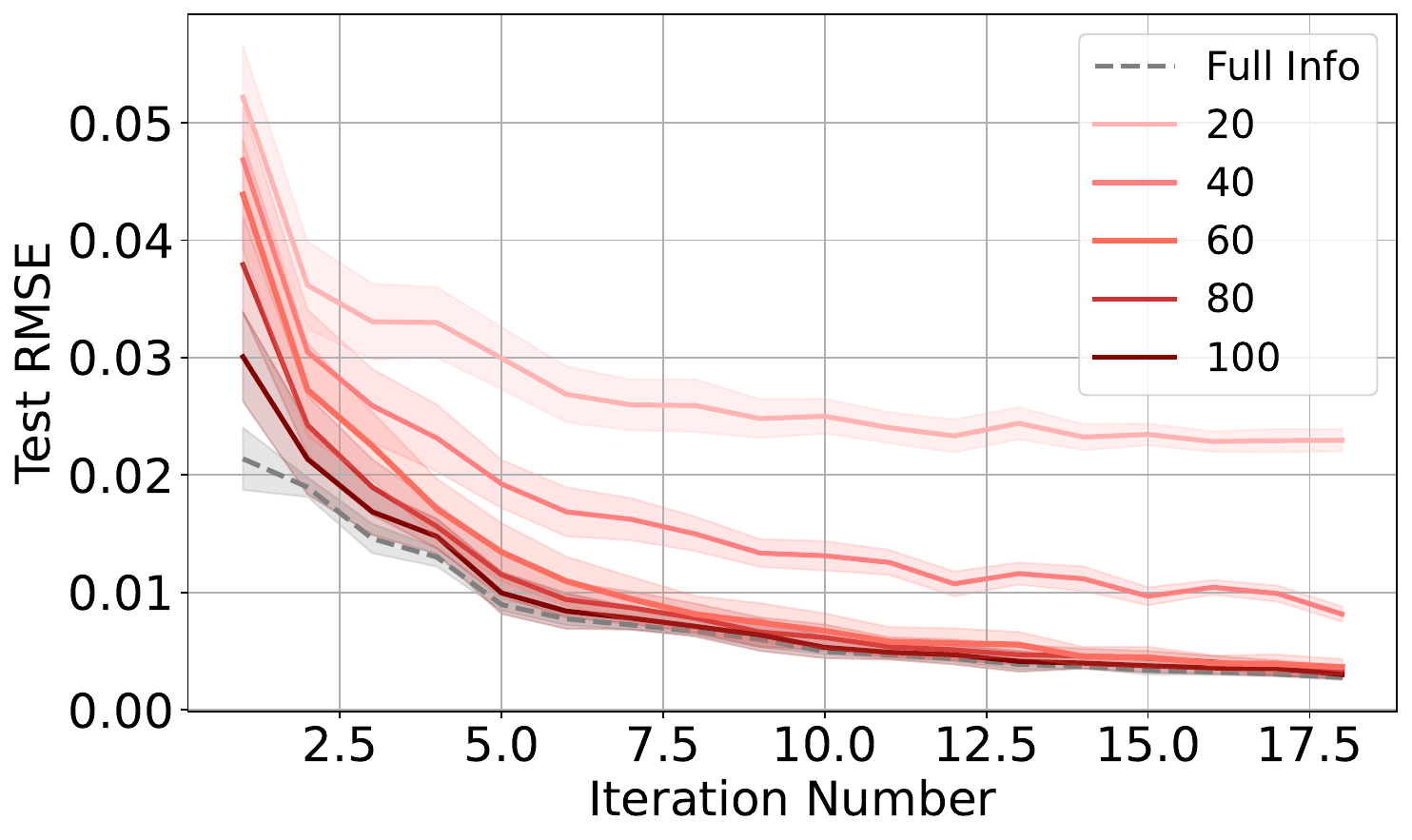}
  \caption{Burgers: effect of varying per-instance query budget $B$ on active learning performance.}
  \label{fig:vary_grid_points}
\end{figure}





For each dataset, we maintained 1,000 training instances and 200 held-out test instances.
From the training set, 100 are reserved for surrogate pretraining and the  900 for active acquisition (18 iterations with 50 instances per iteration). 
Results are averaged over five independent runs with different seeds and constant train/test splits. Our primary metric is RMSE on a dense evaluation grid, and we report mean~$\pm$~std across runs. We compare our \algname against standard heuristic baselines (uniform, random, gradient, variance, intensity), with all methods sharing the same budget, solver query count per iteration, and surrogate retraining cadence. Hyperparameters (networks, optimizers, replay/config for RL, proxy settings, and surrogate schedules) are fixed once chosen and summarized in \appref{sec:exp-supp}. 

To ensure the feasibility of our pipeline, we employ a custom finite-volume solver designed to provide accurate solutions on non-uniform meshes. Specifically, we employ a custom finite-volume solver with an additional geometry augmentation step that inserts virtual midpoints to cap maximum gaps under sparse grid selections (\appref{sec:exp-simulator}). This solver preserves good solution quality even under an irregular mesh grid, while maintaining a monotone cost–complexity relationship (i.e., solve time increases reasonably with the number of queried points). Such a property makes it well-suited for adaptive acquisition: the policy can flexibly select arbitrary grid points without sacrificing correctness or tractability. In practice, we interpolate as needed to guarantee accurate values at queried locations, ensuring that sample complexity is well-controlled and results are comparable across baselines. This differs from classical hierarchical adaptive mesh refinement, where refinements occur on structured strata; instead, our solver supports localized, irregular sampling that integrates naturally with the learning framework. To further strengthen the surrogate, we apply two modifications for training FNOs on non-uniform grids: (1) a masked training strategy, where gradient updates are computed only on selected mesh grid points; and (2) positional embeddings to improve the network’s ability to process geometric information. For fairness, all baselines are adapted to ensure stable interpolation under sparse selections. Full implementation details are provided in \appref{sec:fno-supp} and \appref{sec:exp-simulator}.



\subsection{Main Results}
\label{subsec: res}

Before presenting the main results, we emphasize that all comparisons in this subsection are conducted using an \emph{oracle uniform solver} rather than our non-uniform PDE solver. The rationale is that on highly irregular meshes, interpolation errors can dominate at queried points, making it difficult to disentangle algorithmic differences from solver artifacts. By employing the oracle solver, we ensure that all heuristics and our RL policy are evaluated under identical, noise-free conditions.

Figures~\ref{fig:burgers_main_result}, \ref{fig:darcy_main_result}, and \ref{fig:lorenz_main_result} present the active learning performance of \algname on Burgers, Darcy, and Lorenz--96, respectively, compared with several heuristic baselines. We also plot the full-information curve, which serves as an information-theoretic lower bound on test error. Across all three systems, the RL policy improves faster and attains lower error than heuristic baselines under identical per-instance budgets and retraining cadence, and consistently approaches the full-information curve. The largest gains appear in the early to mid acquisition stages, where the RL error decreases rapidly while heuristic methods plateau. For example, on Burgers, \algname reaches an RMSE of 0.02 by approximately iteration~6, whereas variance, intensity, and gradient-based heuristics require 9--10 iterations and uniform or random selection require 12 or more, corresponding to a $33\%\sim50\%$ reduction in total labeling cost. 
%
Shaded bands indicate that the results are stable: the variability of \algname is comparable to or smaller than that of the stochastic baselines across acquisition iterations. The same qualitative behavior is observed on Darcy and Lorenz--96, despite their very different dynamics, with Lorenz--96 representing a chaotic lattice system rather than a spatially inhomogeneous PDE. Overall, these results support the core claim that treating point selection as a sequential decision problem yields superior accuracy--vs--budget trade-offs compared to fixed heuristics, both for PDEs with spatial structure and for chaotic dynamical systems.

\begin{figure*}[!h]
    \centering
    \begin{subfigure}[t]{0.48\textwidth}
        \centering
        \includegraphics[width=\linewidth]{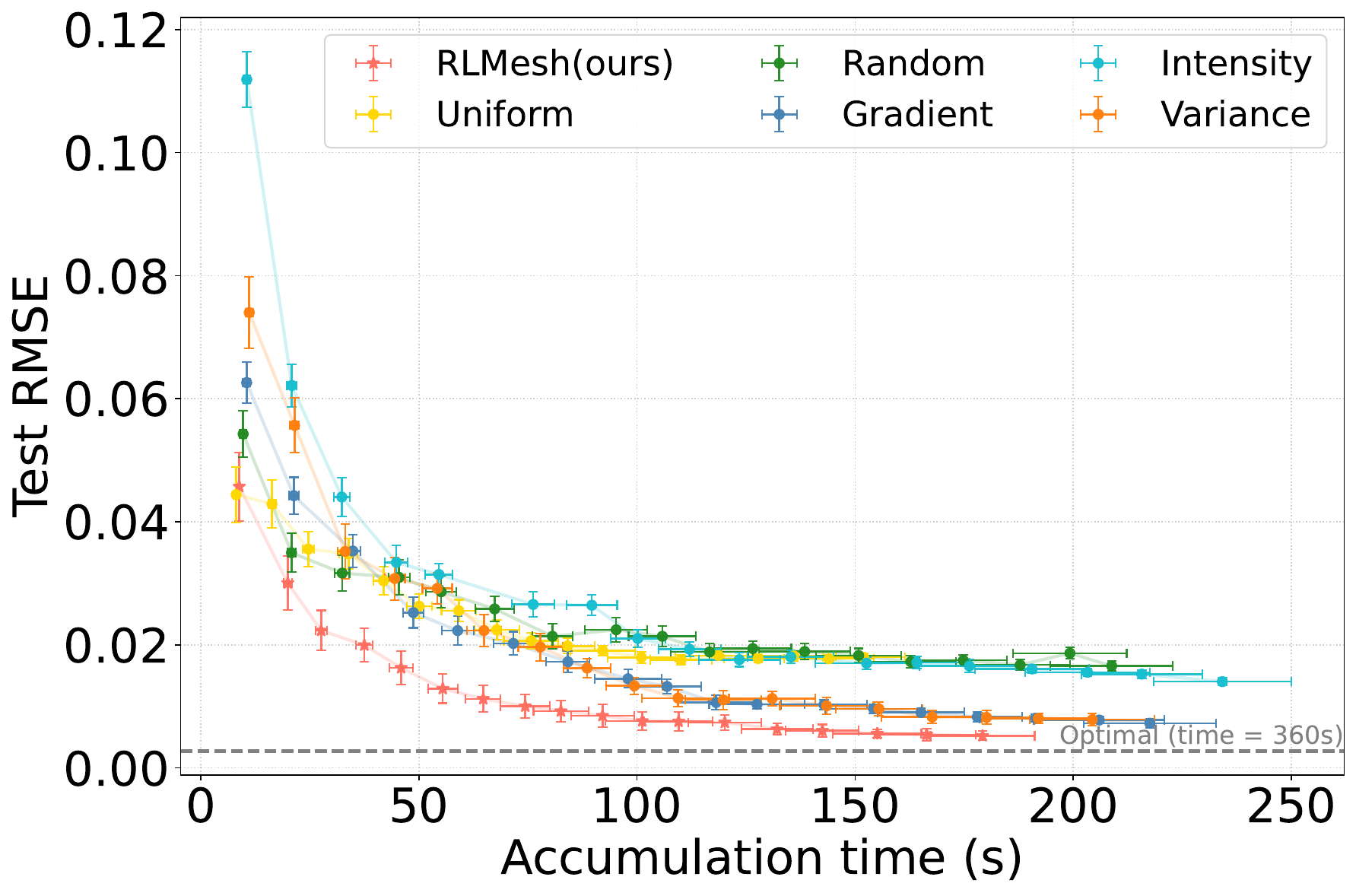}
        \caption{Time–error illustration on Burgers under different selection strategies.
        Each dot corresponds to an acquisition iteration.
        The dashed grey line marks the optimal RMSE and solving time under full information.
        \algname consistently achieves lower error with shorter solving time compared to heuristic baselines.}
        \label{fig:time_error}
    \end{subfigure}\hfill
    \begin{subfigure}[t]{0.48\textwidth}
        \centering
        \includegraphics[width=\linewidth]{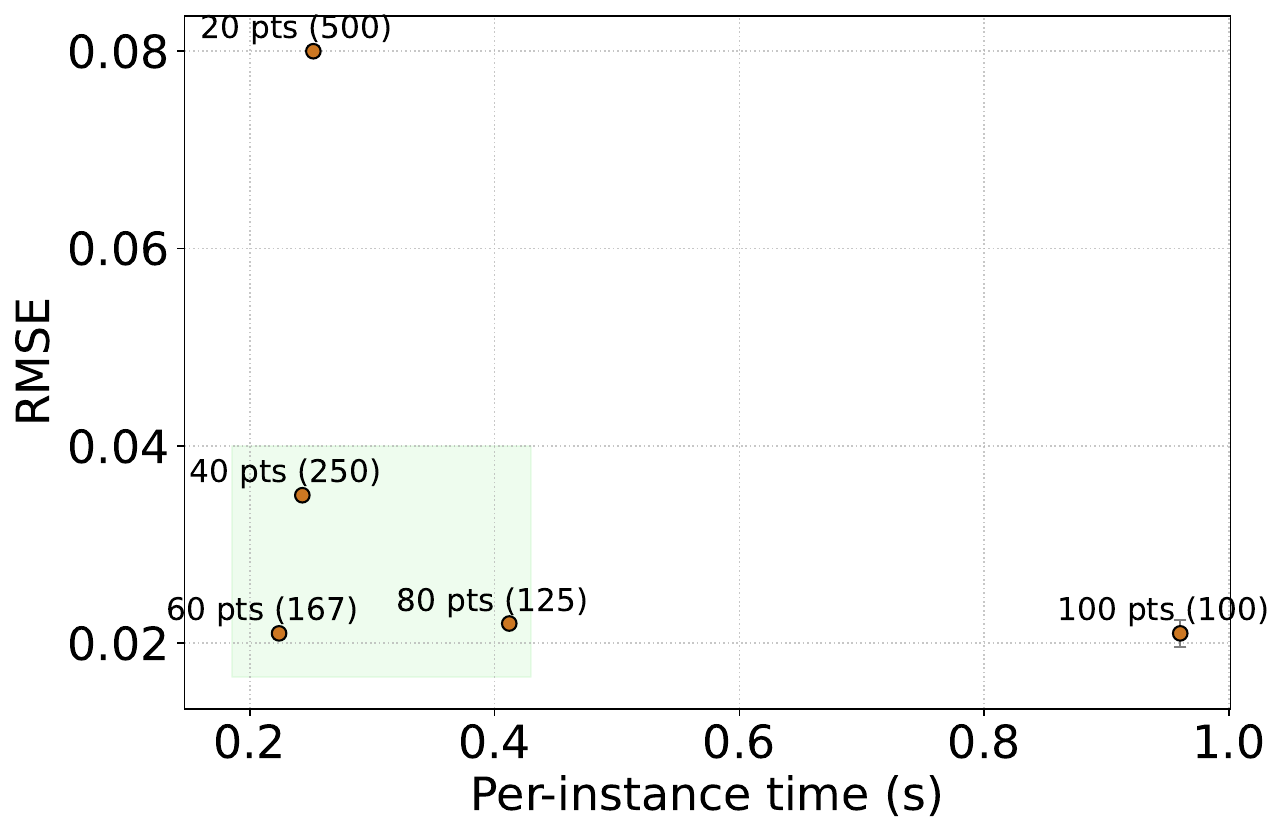}
        \caption{Time–error tradeoff using \algname on Burgers under a fixed total budget.
        Each point corresponds to a different per-instance grid point budget,
        annotated as “number of grid points (number of instances).”
        The shaded green region indicates the preferable regime, balancing low error with shorter solve times.}
        \label{fig:tradeoff}
    \end{subfigure}

    \caption{Time–error analysis on Burgers.
    (a) Comparison across selection strategies.
    (b) Tradeoff across per-instance budgets using \algname.}
    \label{fig:time-tradeoff}
\end{figure*}

\figref{fig:vary_grid_points} shows how performance varies with different per-instance budget $B$. With only 20 or 40 grid points, the RL policy is unable to drive error down substantially—both curves plateau at relatively high RMSE.
To understand where the limitation originates, we ran experiments with alternative heuristics and found it is primarily due to FNO capacity rather than the RL policy, as shown in \appref{sec:fewergrid-exp}. In contrast, once the budget reaches 60 grid points, the loss drops quickly and ultimately converges to the same level as the 80- and 100-grid point settings. This indicates that while additional mesh points (80, 100) yield slightly faster early progress, the marginal returns diminish: all three eventually overlap near the full-information lower bound. The variance bands also become narrower with increasing budget, suggesting more stable learning. Overall, 60 appears to be the most cost-efficient setting, achieving the same accuracy as larger budgets while requiring lower labeling cost.

%% file: sections/analysis.tex
\section{Analysis}
\paragraph{Time-error tradeoff.} In contrast to Section~\ref{subsec: res}, where we isolated algorithmic effects using an oracle solver, here we explicitly account for \emph{simulation time}. Because solver runtime depends strongly on mesh regularity and numerical schemes, we report results using our custom finite-difference solver, which reflects the wall-clock cost of non-uniform selections. We also validate that our custom solver preserves these conclusions under non-uniform selections (\appref{sec:app:custom_simulator}). To avoid ambiguity, we emphasize that each figure uses a single, shared solver across all methods: \figref{fig:main_results_all} evaluates all algorithms under the same oracle uniform-grid solver to isolate accuracy effects, while \figref{fig:time-tradeoff} evaluates all algorithms under the same non-uniform solver to assess time–error tradeoffs.

\begin{figure*}[t]
    \centering
    \begin{subfigure}[t]{0.46\textwidth}
        \centering
        \includegraphics[height=3.5cm]{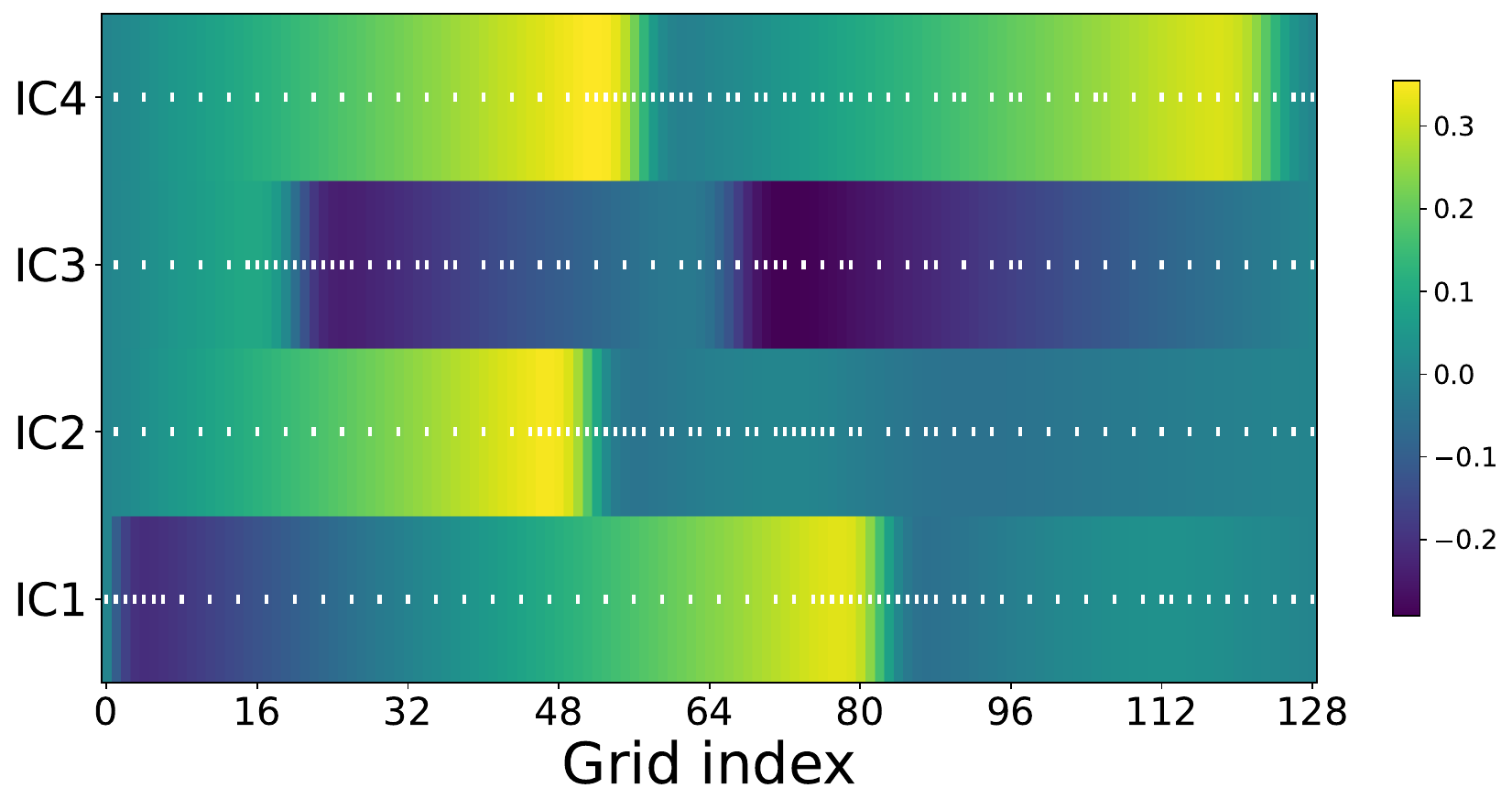}
        \caption{Burgers: grid point selection on four different ICs.}
        \label{fig:burgers_sensor4}
    \end{subfigure}\hfill
    \begin{subfigure}[t]{0.52\textwidth}
        \centering
        \includegraphics[height=3.6cm]{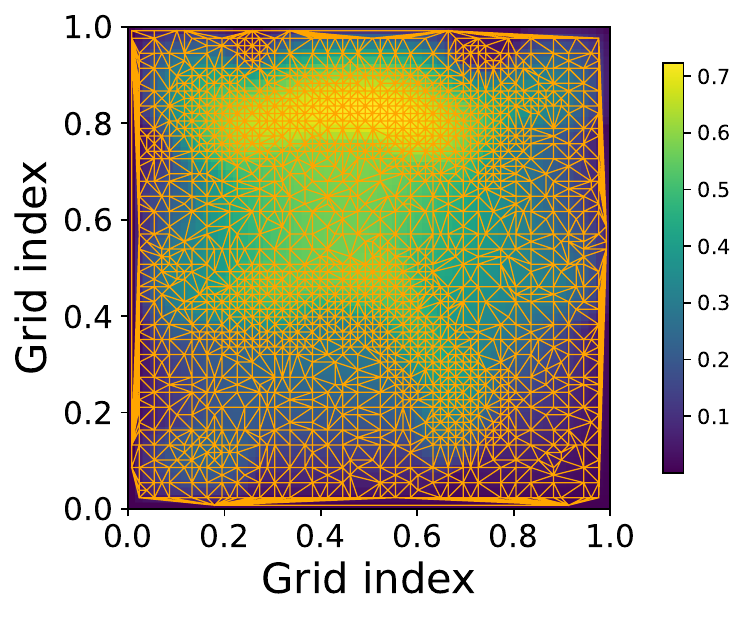}\hfill
        \includegraphics[height=3.6cm]{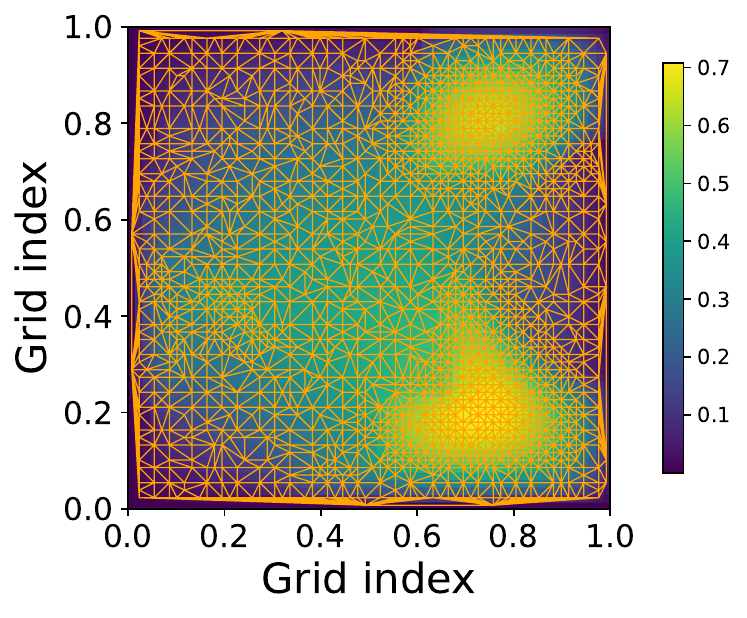}
        \caption{Darcy: grid point selections on two representative input states.}
        \label{fig:darcy_sensor}
    \end{subfigure}

    \caption{Visualization of grid point selections by \algname on Burgers and Darcy. 
    (a) Burgers: selections adapt to steep shock fronts and turning points while maintaining scaffolding in smoother regions. 
    (b) Darcy: selections concentrate along high-contrast channels and near boundary layers, with adaptive coverage in homogeneous regions.}
    \label{fig:grid_selection}
\end{figure*}

\figref{fig:time_error} presents the tradeoff between accumulation time and test error. Across acquisition iterations, \algname consistently reduces error more quickly than heuristic baselines, approaching the full-information lower bound in substantially less time. In particular, to reach the 0.02 error threshold, \algname requires only about 40 seconds of accumulated simulation, whereas variance, intensity, and gradient heuristics require roughly 80–120 seconds, and uniform or random baselines require well over 150 seconds. This gap highlights that \algname not only achieves superior predictive accuracy but also makes far more efficient use of the simulation budget, converting the same amount of wall-clock computation into significantly larger accuracy gains.

In \figref{fig:tradeoff} we fix the total label budget at 10,000 points and vary the per-instance grid point budget which also sets the number of instances (10,000/budget) The scatter reveals a clear Pareto region (lower error at lower per-instance time). Using 20 grid points is faster but underfits (higher error), whereas 100 grid points yields only modest error gains at substantially higher time per initial-condition. We therefore adopt 60 grid points as a practical operating regime for simulator-assisted acquisition.

\begin{figure*}[!th]
  \centering

  \begin{subfigure}[t]{0.4\textwidth}
    \centering
    \includegraphics[height=3cm]{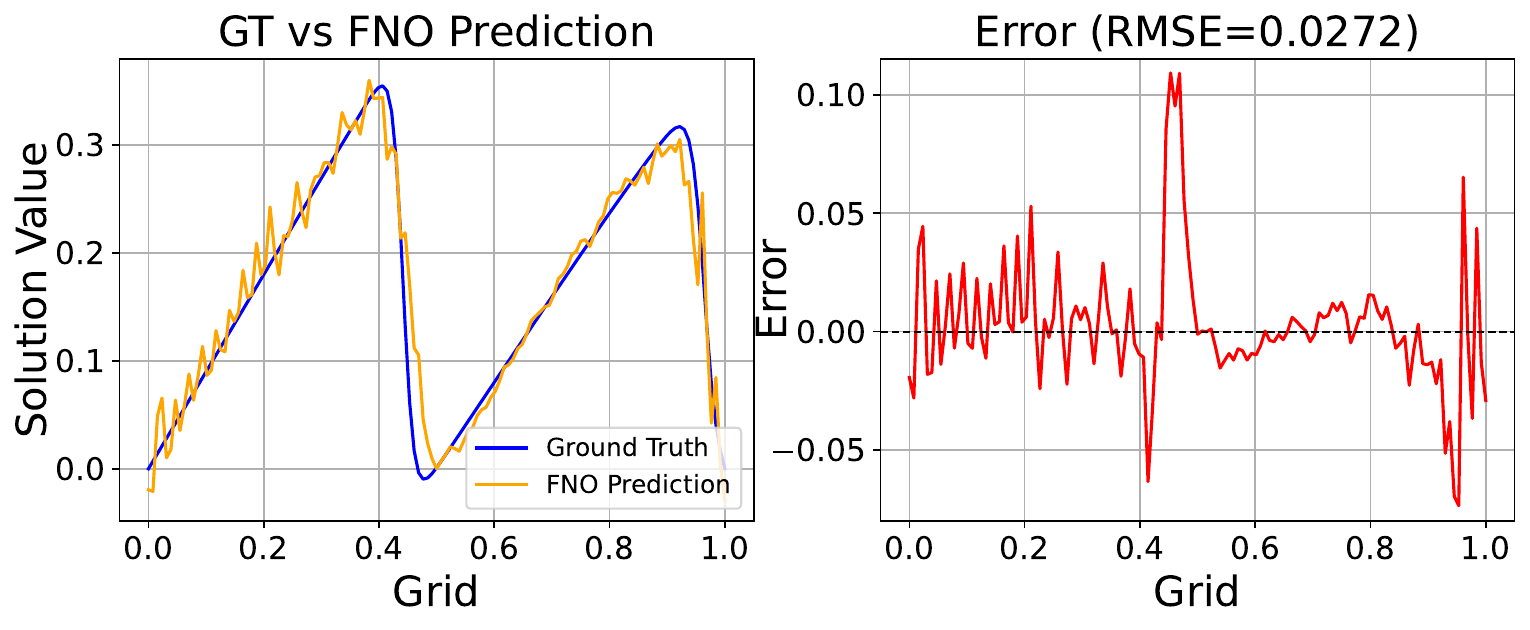}
    \includegraphics[height=3cm]{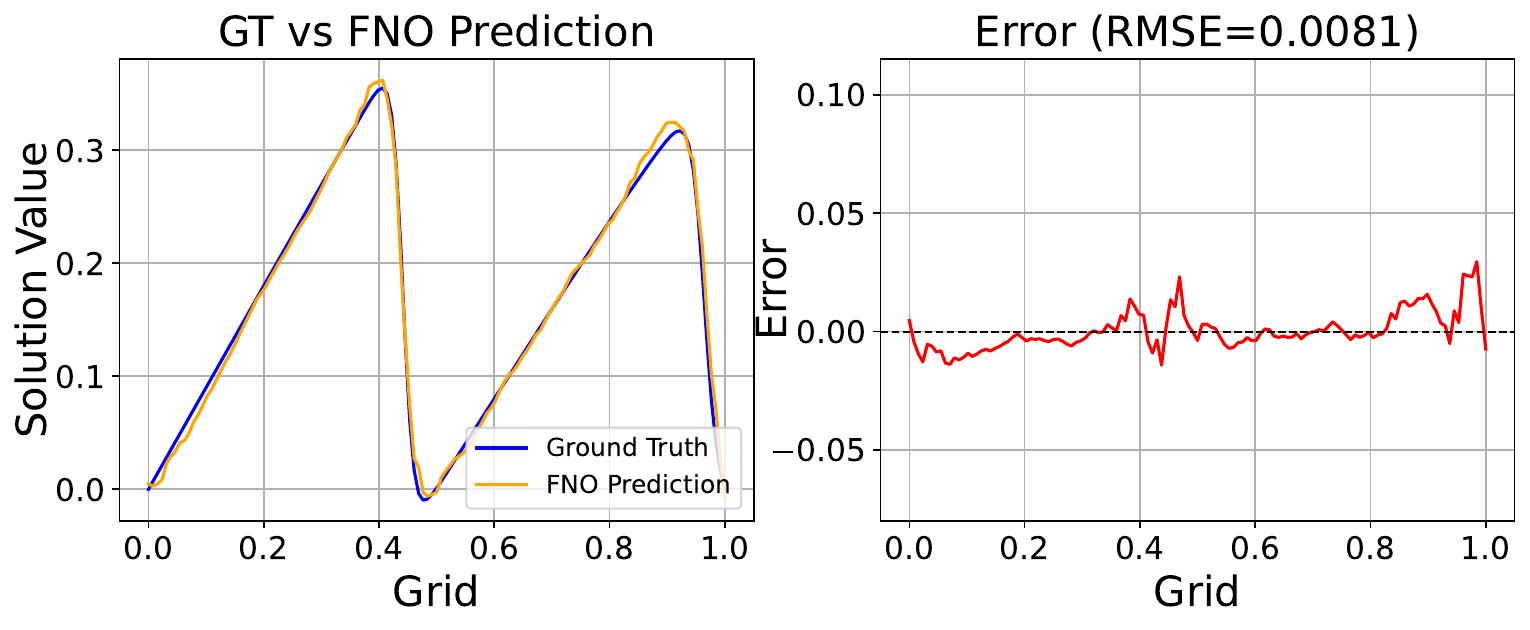}
    \includegraphics[height=3cm]{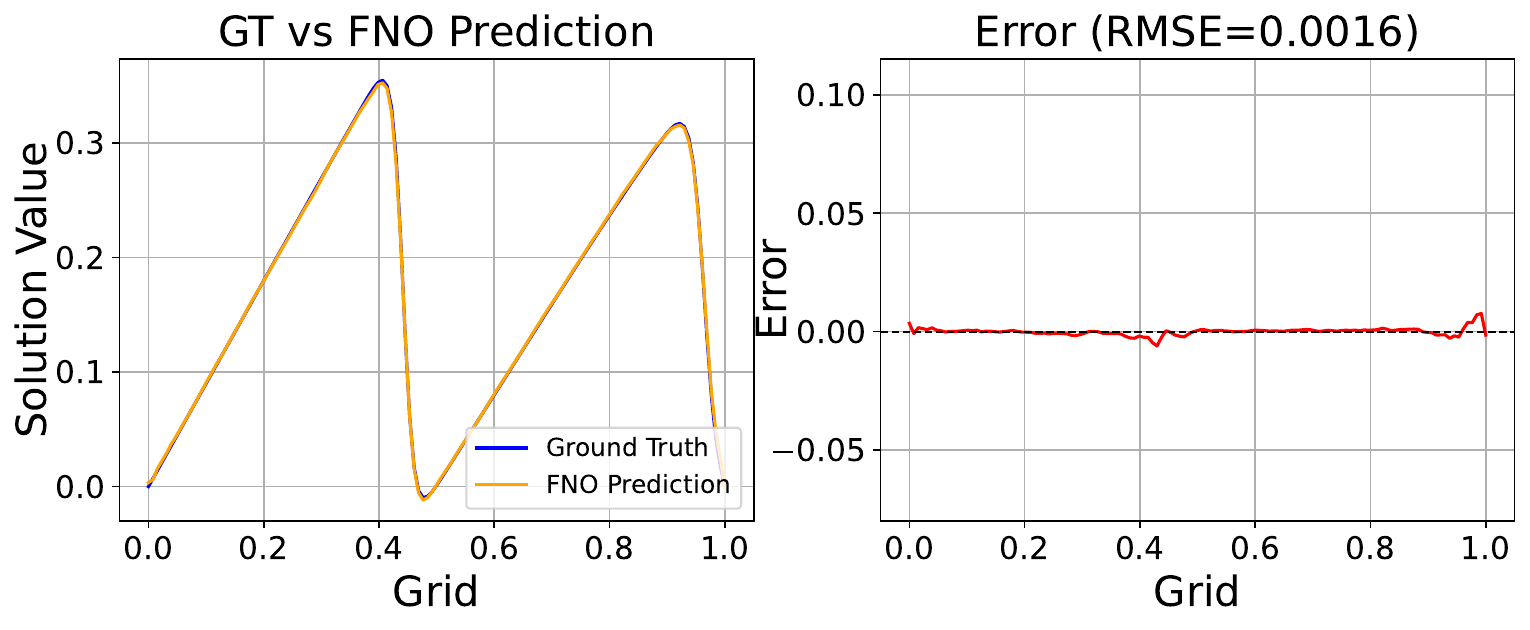}
    \caption{Learning progress on Burgers: pretrained, 100 samples, and 900 samples. 
    Each row shows the ground truth, predictions, and the error map.}
    \label{fig:al_progression_burgers}
  \end{subfigure}\hfill
  \begin{subfigure}[t]{0.55\textwidth}
    \centering
    \includegraphics[height=3cm]{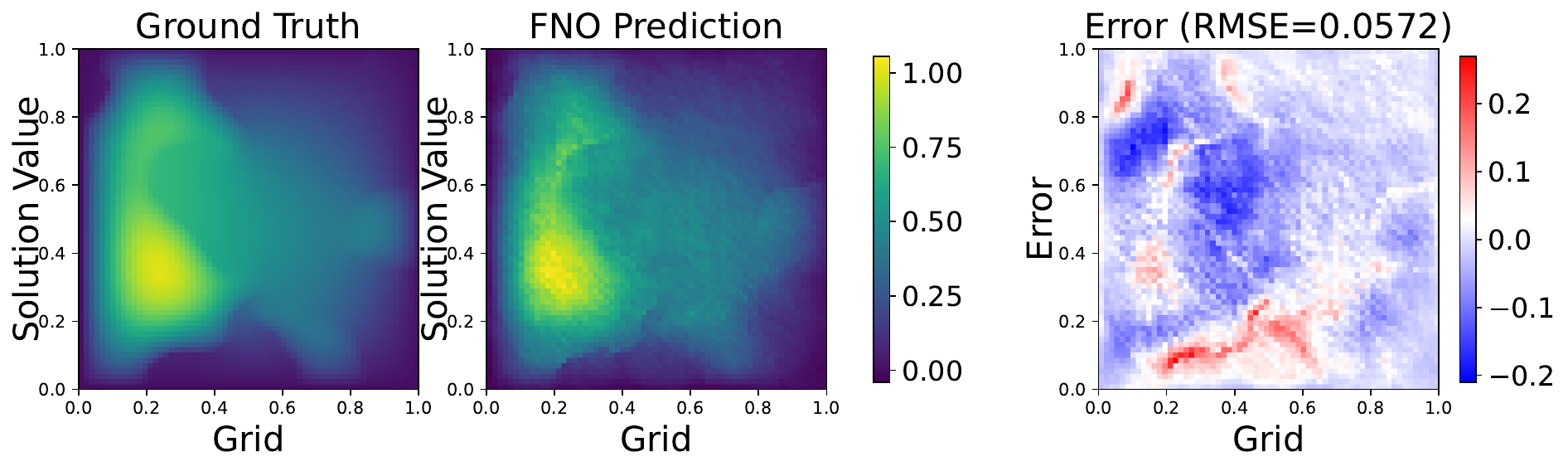}
    \includegraphics[height=3cm]{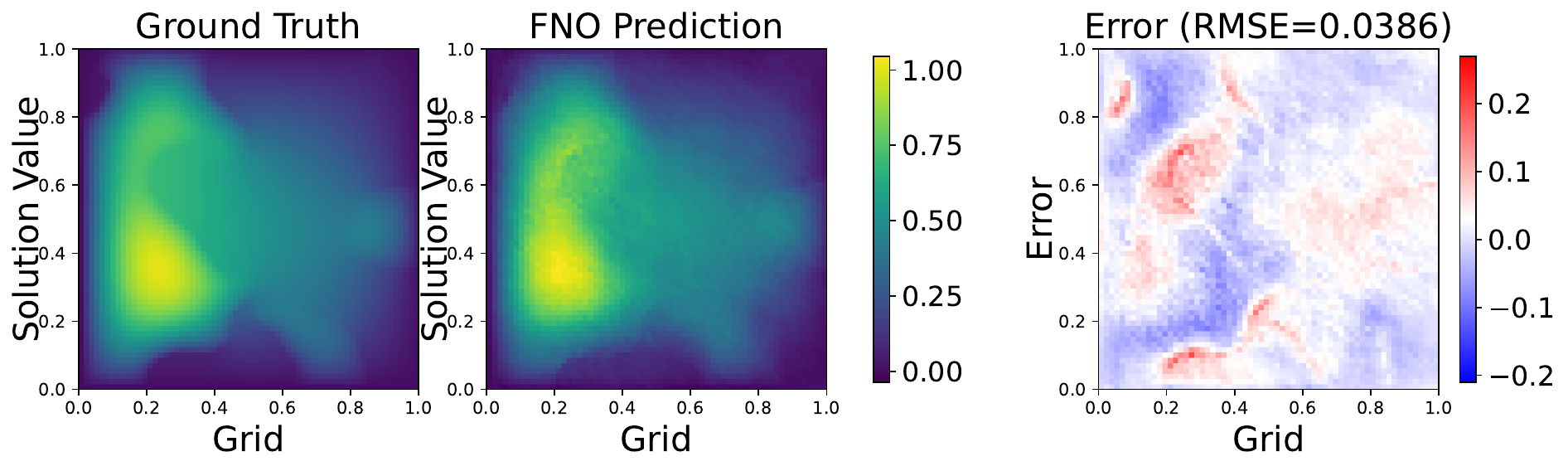}
    \includegraphics[height=3cm]{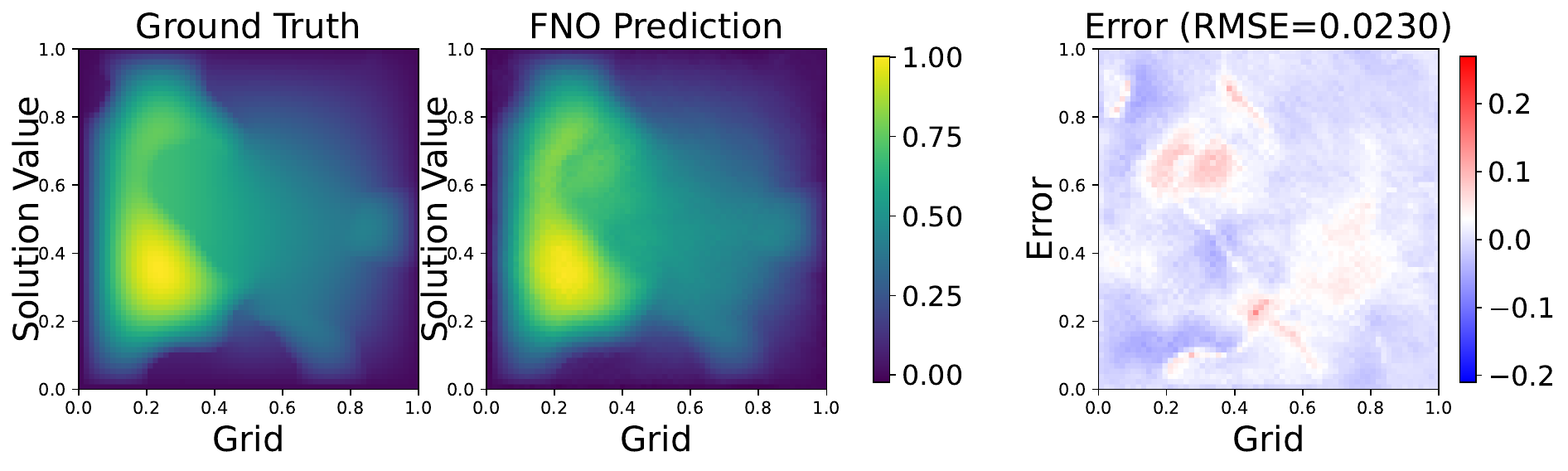}
    \caption{Learning progress on Darcy: pretrained, 100 samples, and 900 samples. 
    Each row shows the ground truth, predictions, and the error map.}
    \label{fig:al_progression_darcy}
  \end{subfigure}

  \caption{Learning progression under active acquisition. 
  (a) Burgers: the model rapidly reduces bias and oscillatory error, with residual discrepancies confined to shock neighborhoods. 
  (b) Darcy: coarse structure is captured early, while later acquisitions refine high-contrast channels and boundary layers.}
  \label{fig:al-progression-main}
\end{figure*}

\paragraph{Qualitative analysis.} We next turn to analyzing the grid point selections learned by our policy. For Burgers in \figref{fig:burgers_sensor4}, across the four instances the learned policy concentrates selection around steep fronts and turning points, while keeping scaffold in smooth ramp regions. For Darcy in \figref{fig:darcy_sensor}, grid points cluster along high-contrast channels and near boundary layers, with coverage in homogeneous areas. The patterns shift across instances, confirming per-instance adaptivity rather than a static mask, and showing that the learned policy adapts naturally to the characteristic features of each PDE family.  We do not include a corresponding visualization for Lorenz--96, as it lacks a continuous spatial geometry and selected indices do not admit the same geometric interpretation as PDE mesh points.

The benefits of such adaptive selection are reflected in the learning curves. 
For Burgers in \figref{fig:al_progression_burgers}, from pretrain to 100 samples the model quickly locks onto the global waveform, eliminating large bias and smoothing the oscillatory error. The remaining error is confined to a narrow neighborhood around the shock/turning points. By 900 samples the curve is visually indistinguishable from ground truth except for tiny boundary artifacts. For Darcy in \figref{fig:al_progression_darcy}, pretrain reconstructions capture coarse structure but miss high-contrast channels; with 100 samples the bulk geometry is recovered while error concentrates along sharp interfaces. By 900 samples the error ridges thin and weaken, especially around permeability contrasts and boundary layers, indicating that later acquisitions act as local refinements rather than global corrections. The spatial patterns of error reduction closely align with our selected mesh grid points, underscoring the informativeness of the chosen grid points.

\paragraph{Mesh-level vs. instance-level adaptivity.}
\begin{table*}[t]
\centering
\caption{Burgers equation: \algname with instance-level baselines (mean $\pm$ std).}
\label{tab:burgers_instance_level}
\begin{tabular}{c|ccc}
\hline
Iteration & RLMesh & Self-MI & LCMD \\
\hline
1  & \textbf{0.0439 $\pm$ 0.0047} & 0.0489 $\pm$ 0.0108 & 0.0464 $\pm$ 0.0169 \\
6  & \textbf{0.0109 $\pm$ 0.0021} & 0.0146 $\pm$ 0.0014 & 0.0126 $\pm$ 0.0007 \\
12 & \textbf{0.00566 $\pm$ 0.00128} & 0.00697 $\pm$ 0.00084 & 0.00618 $\pm$ 0.00041 \\
18 & \textbf{0.00364 $\pm$ 0.00067} & 0.00418 $\pm$ 0.00038 & 0.00384 $\pm$ 0.00010 \\
\hline
\end{tabular}
\end{table*}

Note that \algname was designed for per-instance spatial point selection under a strict local query budget. Meanwhile, prior frameworks such as MRA-FNO \citep{Li2023MRAFNO} and AL4PDE \citep{musekamp2025active} focus on instance-level acquisition (choosing which simulations or global resolutions to run), assuming that each selected instance returns a full field solution. To contextualize \algname relative to such instance-level active learning methods, we adapt acquisition components from MRA-FNO and AL4PDE to our single-fidelity setting. Specifically, we implement a mutual-information–based selector (Self-MI, derived from MRA-FNO) and LCMD (the strongest AL4PDE heuristic for Burgers), while holding the remainder of the pipeline fixed. Since these baselines select entire instances and return full-field solutions, we equalize total query budgets by assigning each selected instance a fixed 60-point uniform grid. Table \ref{tab:burgers_instance_level} reports RMSE under identical total query budgets on Burgers. RLMesh consistently achieves lower error across acquisition iterations, indicating that instance-only selection cannot exploit fine-grained spatial structure that per-instance spatial adaptivity captures.

%% file: sections/conclusion.tex
\section{Conclusion}
We proposed \algname, a co-evolving active learning framework that casts per-instance, budgeted mesh-point selection as a sequential decision problem. By coupling a non-uniform–capable neural operator surrogate with an RL policy and leveraging a lightweight proxy for aligned terminal rewards, our method enables practical online policy improvement under strict labeling budgets. Across Burgers (1D) and Darcy (2D), our method consistently outperforms heuristics in active learning performance. On Burgers, detailed experiments on error–versus–budget and time–error trade-offs further show that targeted acquisition achieves comparable or better surrogate accuracy at substantially lower cost.
Future directions include extending the framework to space–time sensing and multi-fidelity costs, scaling to higher-dimensional or irregular geometries with geometry-aware operators, and exploring joint instance-plus-grid-points selection with theoretical guarantees on sample efficiency. Overall, our results demonstrate that reinforcement learning–driven adaptivity provides a principled and effective path toward cost-efficient PDE surrogate modeling.

%% file: sections/acknowledgment.tex
\section*{Acknowledgment}
This work was partially done when CL was at the University of Chicago. We gratefully acknowledge the support of DOE 0J-60040-0023A, AFOSR FA9550-24-1-0327, NSF IIS 2313131, NSF IIS 2332475, the NSF-Simons AI-Institute for the Sky (SkAI) via grants NSF AST-2421845 and Simons Foundation MPS-AI-00010513, and the University of Chicago’s Research Computing Center for their support of this work.